\documentclass{article}

\PassOptionsToPackage{square,numbers}{natbib}

\usepackage[preprint]{neurips_2025}

\usepackage[utf8]{inputenc} 
\usepackage[T1]{fontenc}    
\usepackage{hyperref}       
\usepackage{amsmath} 
\usepackage[capitalize,nameinlink]{cleveref}
\usepackage{url}            
\usepackage{booktabs}       
\usepackage{amsfonts}       
\usepackage{nicefrac}       
\usepackage{microtype}      
\usepackage{xcolor}         
\usepackage{graphicx} 
\usepackage{subcaption}
\usepackage[dvipsnames]{xcolor}
\usepackage{pifont}
\usepackage[table]{xcolor}
\definecolor{lightgray}{gray}{0.9}
\definecolor{lightorange}{RGB}{231,171,110}
\definecolor{lightgreen}{RGB}{110,189,181}
\definecolor{lightblue}{RGB}{118,159,189}
\definecolor{lightred}{RGB}{225,114,100}
\usepackage{url}
\usepackage{enumitem}
\usepackage[most]{tcolorbox}
\usepackage{listings}
\usepackage{xcolor}
\usepackage{courier} 
\usepackage{arydshln}

\usepackage{amssymb}
\newcommand{\tick}{\ding{51}}%
\newcommand{\cross}{\ding{55}}

\newcommand{\cmark}{\textcolor{ForestGreen}{\ding{51}}}%
\newcommand{\xmark}{\textcolor{red}{\ding{55}}}%

\definecolor{cornflowerblue}{rgb}{0.39, 0.58, 0.93}
\definecolor{tomato}{rgb}{1.0, 0.39, 0.28}
\definecolor{gold}{rgb}{1.0, 0.84, 0.0}
\definecolor{pink}{rgb}{1.0, 0.0, 0.4}

\newcommand{\blueprompt}[1]{\textcolor{blue}{\texttt\textbf{{#1}}}}

\usepackage{makecell}
\usepackage{wrapfig}
\usepackage{amsmath}  
\title{\textit{From Objects to Anywhere}: A Holistic Benchmark for Multi-level Visual Grounding in 3D Scenes}

\author{
  Tianxu Wang$^{1}$\thanks{Equal contribution} \quad \quad
  Zhuofan Zhang$^{1,2}$\footnotemark[1] \quad \quad
  Ziyu Zhu$^{1,2}$ \quad \quad
  Yue Fan$^{1}$ \\
  \textbf{Jing Xiong}$^{1,3}$ \quad \quad
  \textbf{Pengxiang Li}$^{1,4}$ \quad \quad
  \textbf{Xiaojian Ma}$^{1}$ \quad \quad
  \vspace{0.5em}
  \textbf{Qing Li}$^{1}$\thanks{Corresponding author}\\
$^1$State Key Laboratory of General Artificial Intelligence, BIGAI \\
\vspace{0.5em}
$^2$Tsinghua University $^3$Peking University $^4$Beijing Institute of Technology \\
\texttt{\{wangtianxu,liqing\}@bigai.ai}
}

\begin{document}

\maketitle

    \vspace{-2em}
    \centerline{
    \textbf{Project page: \textcolor{lightblue}{\href{https://anywhere-3d.github.io}{https://anywhere-3d.github.io}}}
    }

    \begin{figure}[htbp]
      \centering
        \includegraphics[width=1\linewidth]{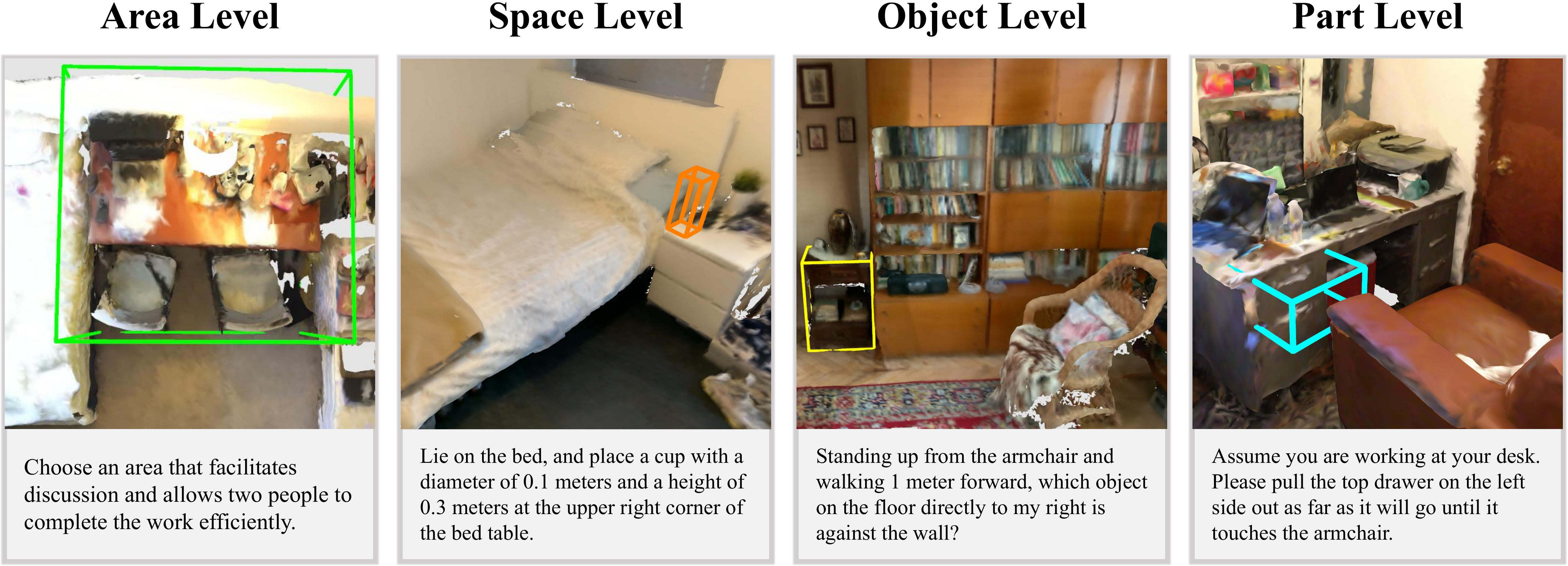}
        \caption{Multi-level Visual grounding in 3D Scenes: area, space, object, and part. Examples illustrate visual grounding of daily life expressions, from functional \textbf{areas} for collaborative study, to placing a cup on a nightstand in unoccupied \textbf{space}, referring to an \textbf{object} via its spatial distance from the armchair, or moving \textbf{part} of an object such as pulling out a drawer from a cabinet.}
        \label{fig:teaser}
    \end{figure}

\begin{abstract}

3D visual grounding has made notable progress in localizing objects within complex 3D scenes. However, grounding referring expressions beyond objects in 3D scenes remains unexplored. In this paper, we introduce \textbf{Anywhere3D-Bench}, a holistic 3D visual grounding benchmark consisting of 2,886 referring expression-3D bounding box pairs spanning four different grounding levels: human-activity \textit{areas}, unoccupied \textit{space} beyond objects, individual \textit{objects} in the scene, and fine-grained object \textit{parts}. We assess a range of state‐of‐the‐art 3D visual grounding methods alongside large language models (LLMs) and multimodal LLMs (MLLMs) on Anywhere3D-Bench. Experimental results reveal that space-level and part-level visual grounding pose the greatest challenges: space‐level tasks require a more comprehensive spatial reasoning ability, for example, modeling distances and spatial relations within 3D space, while part-level tasks demand fine-grained perception of object composition. Even the best-performing models, Google Gemini-2.5-Pro and OpenAI o3, achieve just around 30\% accuracy on space-level tasks and around 40\% on part-level tasks, significantly lower than its performance on area-level and object-level tasks. These findings underscore a critical gap in current models’ capacity to understand and reason about 3D scenes beyond object-level semantics.

\end{abstract}

\section{Introduction}

When instructed to place a floor lamp next to an armchair, humans can visually ground it in the scene, estimating its base diameter and height, imagining its precise alignment with the armchair, and judging whether it fits naturally within the 3D environment. Humans can naturally perceive, reason about, and localize expressions to ``anywhere'' in 3D scenes. Yet can today’s 3D vision–language models ground free-form referring expressions to precise positions and dimensions in a 3D scene, especially when those expressions refer to regions beyond objects?

Existing 3D visual grounding models, pretrained on large 3D scene datasets, excel at aligning expressions to objects in a scene~\cite{scanrefer, multi3drefer, scanqa, scanreason, SG3D, huang2025unveiling}. However, these models remain constrained to object-level alignment, with limited attention paid to the broader spatial regions beyond objects. Meanwhile, with the rapid development of Multimodal Large Language Models (MLLMs), an increasing number of studies have begun to explore their ability to perceive and reason about spatial intelligence from 2D images or videos~\cite{spatialrgpt, spatialvlm, thinkinginspace, orientanything, seeingfromanotherperspective,videoagent,fan2025embodied,zhu2025mtu}. However, their ability to predict the positions and sizes of 3D bounding boxes corresponding to free-form referring expressions anywhere in 3D space, including both objects and regions beyond object boundaries, remains largely unexplored.

\begin{figure}[htbp]
        \centering
        \includegraphics[width=1\linewidth]{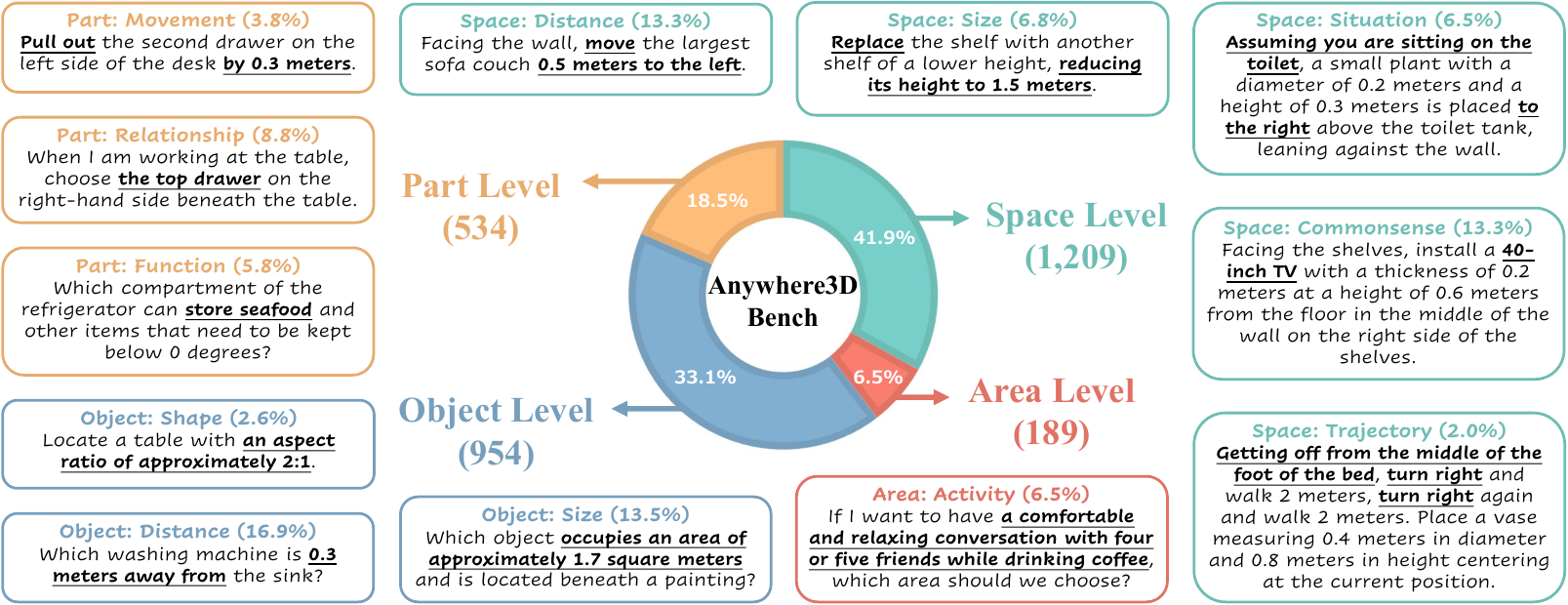}
        \caption{Multi-level visual grounding (\textcolor{lightorange}{Part}, \textcolor{lightgreen}{Space}, \textcolor{lightblue}{Object}, \textcolor{lightred}{Area}) with distinct expression types. \textbf{\underline{Emphasized segments}} highlight phrases aligned with their respective expression type.}
        \label{fig:data_stats}
\end{figure}

To bridge this gap, we introduce \textit{Anywhere3D-Bench}, a holistic benchmark with 2,886 referring expression-3D bounding box pairs, categorized into four visual grounding levels: area, space, object, and part, as illustrated in \cref{fig:teaser}. To the best of our knowledge, we are the first to propose a 3D visual grounding benchmark that spans four hierarchical levels of grounding granularity, particularly on aligning expressions with 3D locations and sizes at \textbf{space level}. More diverse and illustrative examples can be found in \cref{fig:data_stats}. At each level, we design distinct types of referring expressions to evaluate models’ abilities to perceive and reason about various aspects of 3D scene. 

We conduct experiments on three categories of models
on \textit{Anywhere3D-Bench}: (1) LLMs with textual inputs , (2) MLLMs with both visual and textual inputs, and (3) 3D visual grounding specialist models. Evaluation results reveal that current models perform poorly on our benchmark, particularly on space-level and part-level tasks. Space-level tasks require modeling spatial relationships and distances in unoccupied space beyond individual objects, while part-level tasks demand first identifying the relevant object and then reasoning over its fine-grained structure to predict the appropriate bounding box size and position. 
Among all models, Google Gemini-2.5-pro and OpenAI o3---strong MLLMs with visual reasoning capabilities---achieve the best performance on Anywhere3D-Bench, yet both still record around 30\% accuracy on space-level and around 40\% on part-level tasks. These are significantly lower than their performance on the object-level(62.37\% on average) and area-level tasks(85.72\% on average).

With visual inputs from video frames of the scene as well as bird’s-eye view image, MLLMs outperform their non-visual LLM counterparts, particularly on object-level and part-level tasks, where detailed visual cues, such as object appearance and structure, can be leveraged. In contrast, gains at the space level are limited, suggesting that spatial relational reasoning in 3D space remains a significant bottleneck for MLLMs. Notably, LLMs and MLLMs generally outperform 3D visual grounding specialist models, especially on space-level tasks. This advantage can be attributed to their pretraining on large-scale image-text datasets, which endows them with stronger perceptual and understanding capabilities. Furthermore, their exposure to real-world knowledge enables a degree of commonsense reasoning, allowing them to infer 3D locations even beyond objects in the scene.

At the object level, our benchmark specifically assesses models’ ability to understand quantitative object sizes and inter-object distances, which are rarely emphasized in previous 3D visual grounding benchmarks. The best-performing 3D visual grounding model, Chat-Scene~\cite{chatscene}, achieves only 31.73\% accuracy on the object-level task, substantially lower than the over 50\% accuracy reported on benchmarks such as ScanRefer~\cite{scanrefer}, highlighting current 3D visual grounding models' limitations in reasoning about precise object dimensions and spatial distance.

To identify the performance bottlenecks of existing MLLMs on our benchmark, we conduct a human-annotated error analysis on Gemini-2.5-Pro, which provides its reasoning process. All errors are categorized into three types: visual perception errors, caused by misrecognition of object labels and attributes; linguistic reasoning errors, stemming from incorrect logical inference; and spatial reasoning errors, resulting from misalignment in spatial relationships. We observe that errors in space-level tasks are primarily due to challenges in linguistic and spatial reasoning, where understanding spatial orientation and performing logical inference are critical. In contrast, errors in part-level tasks are mainly caused by visual perception issues, which require fine-grained recognition of object details. Additionally, we present a qualitative comparison between the best-performing thinking model (i.e., Gemini-2.5-Pro) and a non-thinking model (i.e., GPT-4.1), demonstrating that the thinking model exhibits superior understanding of object orientations and spatial relationships by better leveraging visual inputs.

To summarize, our contributions are as follows:
\begin{itemize}[leftmargin=10pt]
    \item We introduce \textit{Anywhere3D-Bench}, the first benchmark for multi-level 3D visual grounding that extends beyond the object level to cover four grounding levels in 3D scenes: area, space, object, and part.
    \item Experiments on \textit{Anywhere3D-Bench} reveal that space-level and part-level visual grounding are the most challenging tasks. Even the best-performing models, Gemini-2.5-pro and o3 with visual reasoning ability, struggles with these two tasks. Furthermore, compared to MLLMs, 3D visual grounding specialist models exhibit limited performance and poor generalization to space-level tasks.

\end{itemize}

\section{Anywhere3D Benchmark}

\begin{table}[htbp]
  \centering
  \caption{Comparison of Anywhere3D with existing visual grounding benchmarks (test splits). Anywhere3D expands grounding level to \textbf{area}, \textbf{space}, \textbf{object}, and \textbf{part}.}
  \vspace{0.5em}
  \label{tab:dataset_comparison}
  \resizebox{0.9\linewidth}{!}{
  \begin{tabular}{lcccccccc}
  \toprule
  Benchmark  & Area & Space & Object & Part & Test Source & Quality Check & \# Scene & \# Expression \\ \midrule
  ScanRefer~\citep{scanrefer}  & \xmark & \xmark & \cmark & \xmark & Human    & \cmark & 97 & 5,410 \\
  Nr3D~\citep{referit3d}       & \xmark & \xmark & \cmark & \xmark & Human    & \cmark & 130 & 7,805 \\
  Sr3D~\citep{referit3d}       & \xmark & \xmark & \cmark & \xmark & Template & \cmark & 255 & 17,726 \\
  MMScan~\citep{mmscan} & \cmark & \xmark & \cmark & \xmark & GPT-4    & \cmark & 702 & 19,696 \\
  SceneFun3D~\citep{scenefun3d} & \xmark & \xmark & \xmark & \cmark & Human \& Rephrasing & \cmark & 85 & 1,265 \\
  ScanReason~\citep{scanreason} & \xmark & \xmark & \cmark & \xmark & GPT-4    & \cmark & - & 1,474 \\
  \midrule
  \rowcolor{lightgray} \textbf{Anywhere3D (ours)} & \cmark & \cmark & \cmark & \cmark & GPT-4 & \cmark & 276 & 2,886 \\ 
  \bottomrule
  \end{tabular}
  }
\end{table}

\begin{figure}[htbp]
  \centering
  \includegraphics[width=0.85\linewidth]{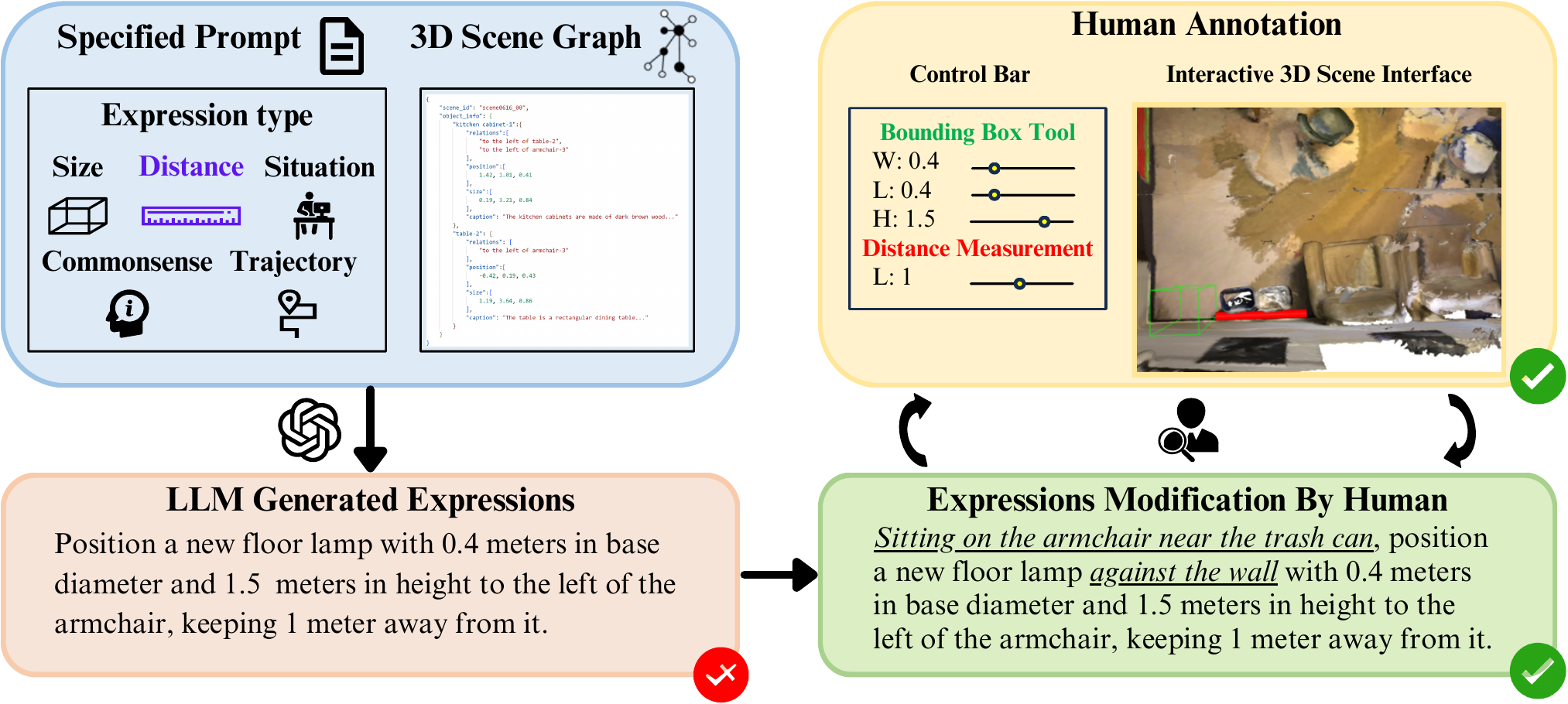}
  \caption{Data generation pipeline of Anywhere3D-Bench. We design specific prompts for different types of expression in four grounding levels. Human annotators are required to annotate the 3D bounding box and refine the GPT-generated expression until they precisely match.}
  \label{fig:data_generation_pipeline}
  \vspace{-1.5em}
\end{figure}
    
      As presented in \cref{tab:dataset_comparison}, we introduce \textit{Anywhere3D-Bench}, which consists of 2,886 referring expression-3D bounding box pairs derived from 276 scenes from the validation sets of ScanNet~\cite{scannet}, MultiScan~\cite{multiscan}, 3RScan~\cite{3RScan}, and ARKitScenes~\cite{arkitscenes}. Inspired by how people refer to 3D scenes in everyday scenarios, we design four levels of grounding granularity and generate referring expressions specifically tailored to each level: Area Level (189), Space Level (1,209), Object Level (954), and Part Level (534). At each level, we design distinct types of referring expressions aiming at evaluating the models' comprehensive capabilities, as further elaborated in the following section.
    
\subsection{Multi-level Visual Grounding}

     We present the data distribution of \textit{Anywhere3D-Bench} in \cref{fig:data_stats}, along with the representative examples of different types of referring expressions for each level. For detailed benchmark data analysis, please refer to Appendix \cref{appendix A: data statistics}.

    \paragraph{Area Level}
     Expressions belonging to the area level typically describe an indoor \textbf{\textit{Activity}}, which requires the model to infer the related functional area composed of multiple objects and the space between them.

   \paragraph{Space Level}
   Expressions in space level refer to spatial regions beyond objects in the 3D scene and are categorized into the following five types: \textbf{\textit{Size}}: Expressions that require directly adjusting the size of an object or performing transformations on its dimension (e.g., length, width, height). This category evaluates models' ability to interpret and manipulate quantitative object dimensions in 3D space. \textbf{\textit{Distance}}:  Expressions involving the relocation of an object or the placement of a new object at a specified distance from another. These tasks test models' ability to reason about quantitative spatial distances and relationships. \textbf{\textit{Situation}}: Expressions involve imagining a scenario in a 3D scene from an egocentric perspective, as introduced in SQA3D~\cite{SQA3d}. This category evaluates models’ ability to understand the situation context and perspective within 3D environments. \textbf{\textit{Commonsense}}: Expressions that include commonsense knowledge about object size (e.g., ``40-inch TV'') or typical spatial locations in a scene (e.g., ``room corner''). \textbf{\textit{Trajectory}}: Expressions that specify the starting point and path of a trajectory, requiring the model to place an object at the endpoint of the trajectory and predict the location and size of the object.

   \paragraph{Object Level} Expressions refer to objects in the 3D scene, following the same setting as in prior works, such as ScanRefer and Nr3D~\cite{scanrefer, referit3d}. However, we place particular emphasis on models' ability in reasoning about the \underline{quantitative} understanding of object \textbf{\textit{Size}},  \textbf{\textit{Shape}}, and inter-object \textbf{\textit{Distance}}. 

    \paragraph{Part Level} 
    Expressions refer to specific parts of objects in a 3D scene and can be categorized into the following three types: \textbf{\textit{Movement}} requires models to predict the bounding box of an object part after certain movement, while \textbf{\textit{Relationship}} and \textbf{\textit{Function}} require models to predict a specific part of an object based on its spatial relationship or function.

\subsection{Data Generation Pipeline}

   As shown in \cref{fig:data_generation_pipeline}, the data generation pipeline of \textit{Anywhere3D-Bench} involves referring expression generation using LLM guided by human-written prompts, along with iterative human annotation and verification. Notably, our annotation interface supports resizing and moving 3D bounding boxes, as well as a distance measurement tool, together enabling precise annotation anywhere in the 3D space.
    
    \paragraph{Referring Expressions Generation}
    To enhance the diversity of referring expressions across various 3D scenes, we leverage GPT-4o~\cite{gpt4o} to generate expressions regarding different grounding levels as well as different types of expressions in each level. For each scene, we first generate a 3D scene graph following SceneVerse~\cite{sceneverse}. Each scene graph contains ground-truth object labels, IDs, and 3D bounding boxes of the objects, as well as object captions and inter-object relationships. We then prompt GPT-4o to generate referring expressions by providing the scene graph along with human-written prompts corresponding to a particular expression type and grounding level.
        
    
    

   \paragraph{Human Annotation and Verification}
   
  To ensure the quality of the benchmark, we construct a human-in-the-loop annotation and verification workflow.
   Annotators are provided with visualizations of the 3D scene as well as ground-truth object labels and sizes, which is adapted from ScanRefer's annotation design.  They are allowed to revise the referring expressions and are required to annotate the corresponding 3D bounding boxes via an interactive interface equipped with a bounding box editor and distance measurement tool. A key requirement is emphasized throughout the workflow: \textbf{Each referring expression must be grounded exactly to one target 3D bounding box in the scene without ambiguity.}
    
    All annotated expressions and 3D bounding boxes are subsequently verified by humans. Any annotation that does not meet the quality criteria is rejected and iteratively revised until it fully complies with the requirements. Please refer to Appendix \cref{appendix A: data generation details} and \cref{appendix A: human annotation and verification details} for additional information.



\section{Experiments and Results}

\subsection{Experimental Setting}
\label{Experimental Setting}
\paragraph{Evaluation Metric}
    In general, we adopt $\mathrm{Acc}@\mathit{k}\mathrm{IoU}$ as the evaluation metric following the standard setting of 3D visual grounding, where IoU is the Intersection over Union between the predicted 3D bounding box and the ground-truth bounding box formatted as $[\mathit{center}_{x}, \mathit{center}_{y}, \mathit{center}_{z}, \mathit{size}_{x}, \mathit{size}_{y}, \mathit{size}_{z}]$. To handle geometric ambiguities at multi-level visual grounding, we apply the following \cref{eq:iou} for IoU computation.
    In our main paper, we set the threshold $t=0.05(m)$, $\mathit{k} = 0.25$ and report $\mathrm{Acc}@0.25\mathrm{IoU}$. For evaluations under other $k$ thresholds and explanation of the IoU formulation, please refer to the Appendix \cref{appendix B: evaluation metrics}.

   \begin{equation}
    \vspace{-0.5em}
    \label{eq:iou}
    \mathrm{IoU} =
    \begin{cases}
    \mathrm{IoU}^{2D}_{xy}, & \text{if } \mathrm{level} = \text{``area''} \\[6pt]
    \mathrm{IoU}^{2D}_{\setminus i} \cdot \mathbf{1}_{\left\{
    |\mathrm{center}_i^{\mathrm{gt}} - \mathrm{center}_i^{\mathrm{pred}}| < t \;\land\;
    \mathrm{size}_i^{\mathrm{pred}} < t
    \right\}}, &
    \parbox[t]{0.45\textwidth}{
    \raggedright
    if $\mathrm{level} \neq \text{``area''}$, \\
    $\mathrm{size}_i^{\mathrm{gt}} < t$, $i \in \{x, y, z\}$
    } \\[6pt]
    \mathrm{IoU}^{3D}, & \text{otherwise}
    \end{cases}
    \end{equation}


\paragraph{Baselines} We evaluate three different types of models on our benchmark:
\begin{itemize}[leftmargin=10pt]
\item \textbf{LLMs:}
    For each expression, the textual scene representation of LLMs are formatted as a scene graph, consisting of ground-truth locations and sizes of objects, as well as object captions. The object captions are generated using Qwen2.5-VL-72B~\cite{qwen2.5-vl} conditioned on multiple object images and a guided captioning instruction (see Appendix \cref{appendix B: Object Caption Generation by qwen2.5-vl} for comprehensive descriptions). 
    Closed-source models, including non-thinking model GPT-4.1~\cite{gpt4.1} and thinking model o4-mini~\cite{o3o4mini}, and open-source models, including non-thinking (Qwen2.5 series ~\cite{qwen2.5, qwen2.5-vl}, DeepSeek-V3-671B\cite{deepseekv3}) and thinking models (Qwen3~\cite{qwen3}, DeepSeek-R1-671B~\cite{deepseekr1}) are benchmarked. 

\item \textbf{MLLMs:}
    Following the setting in GPT4Scene~\cite{gpt4scene}, we incorporate a bird’s-eye view (BEV) image and eight uniformly sampled video frames with object markers as visual inputs, in addition to the textual scene representation used in the LLM setting.
    Closed-source models, including non-thinking GPT-4.1 and thinking model o4-mini, o3 and Gemini-2.5-Flash\cite{gemini-2.5-flash}, Gemini-2.5-Pro\cite{gemini-2.5-pro}, as well as open-source models, including LLaVA-OneVision~\cite{llava-onevision}, Qwen2.5-VL~\cite{qwen2.5-vl}, InternVL3~\cite{internvl3} and GPT4Scene~\cite{gpt4scene} are evaluated. 

\item \textbf{3D Visual Grounding Models:}
    For 3D VG models, We use the ground-truth object mask as input to ensure alignment with the ground-truth object location and size used in both the LLM and MLLM settings, whenever applicable. We evaluate four state-of-the-art specialized 3D visual grounding models: 3D-VisTA~\cite{3dvista}, PQ3D~\cite{pq3d}, Chat-Scene~\cite{chatscene} and Grounded 3D-LLM~\cite{grounded3dllm}. Since Chat-Scene and Grounded 3D-LLM do not provide 3D features for datasets other than ScanNet, their evaluations are limited to the ScanNet portion of our benchmark.
\end{itemize}

Thorough experimental settings and implementations of baselines can be founded in Appendix \cref{appendix B: Baseline Settings}.
\paragraph{Human Evaluation}
   We construct a human evaluation subset of 200 expressions through stratified random sampling across four levels to maintain their original distribution.
   Human evaluators are instructed to annotate the corresponding 3D bounding boxes for each expression. Evaluation results on this subset are reported using the same metric as mentioned above.

\subsection{Main Results}

\begin{table}[htbp]
\vspace{-1.0em}
\caption{Results are presented in $\mathrm{Acc}@0.25\mathrm{IoU}$ on Anywhere3D-Bench. \textit{object bbox} in the table denotes the ground-truth object locations and sizes for simplicity. Chat-Scene*, Grounded 3D-LLM*: evaluated only on ScanNet. Human**: performance evaluated on a subset of 200 expressions across four levels. The best performance in each setting is highlighted in \textbf{bold}, and the second-best is indicated with \underline{underline}.}
\vspace{0.3em}
\renewcommand{\arraystretch}{1.1}
\resizebox{\linewidth}{!}
{

    \begin{tabular}{l c c c c c c}   
    \toprule

    & Open Source 
    & Area Level 
    & Space Level 
    & Object Level 
    & Part Level 
    & Overall \\
    \midrule

    \rowcolor{lightgray}
    \multicolumn{7}{l}{\textbf{LLMs: \textit{object bbox, captions}}}\\
    \multicolumn{7}{l}{\textbf{non-thinking}}\\
    GPT-4.1-2025-04-14          & \cross   & \textbf{76.19}  & \underline{17.28}  & \underline{48.00}  & \underline{22.94}  & \underline{32.34}  \\
    Qwen3-32B(non-thinking)    & \tick   & 54.67 & 9.60  & 31.97  & 12.24  & 20.43  \\
    Qwen2.5-72B      & \tick    & 60.14  &  7.85  & 33.30  & 8.99  & 19.90  \\
    Qwen2.5-VL-72B   & \tick    & 56.35  &  6.87  & 29.19  & 9.93  & 18.05 \\
    DeepSeek-V3-671B-2024-12-26 & \tick    & 61.38  & 9.81 & 41.06  & 15.61  & 24.59 \\
    \hdashline
    \multicolumn{7}{l}{\textbf{thinking}}\\
    o4-mini-2025-04-16          & \cross   & \underline{71.96} & \textbf{18.03} & \textbf{48.69} & \textbf{23.97} & \textbf{32.80}\\
    Qwen3-32B(thinking)        & \tick    & 59.79 & 12.57 & 40.18 & 16.48 & 25.51 \\
    DeepSeek-R1-671B-2025-01-28 & \tick    & \underline{71.96} & 14.61 & 47.76 & 20.92 & 30.49 \\
    \midrule

    \rowcolor{lightgray}
    \multicolumn{7}{l}{\textbf{MLLMs: \textit{object bbox, captions, BEV, video frames}}}\\
    \multicolumn{7}{l}{\textbf{non-thinking}}\\
    GPT-4.1-2025-04-14          & \cross   & 81.48 & 19.03 & 53.88 & 25.85 & 35.90 \\
    Gemini-2.0-Flash(non-thinking)  & \cross   & 68.43 & 13.18 & 45.39 & 19.97 & 28.70\\
    LLaVA-NeXT-Interleave-7B & \tick    & 6.88 &  0.83 &  4.61 &  2.06 &  2.70 \\
    LLaVA-OneVision-7B & \tick    & 19.58 &  2.32 &  8.81 &  4.12 &  5.93 \\
    InternVL3-8B & \tick    & 33.16 &  4.60 &  18.69 & 6.93 &  11.56 \\
    Qwen2.5-VL-72B   & \tick    & 57.16 &  10.56 & 40.74 & 13.80 & 24.19 \\
    GPT4Scene\phantom{X} & \tick    & 15.34 &  7.19 & 25.16 & 11.99 & 14.55 \\
    \hdashline
    \multicolumn{7}{l}{\textbf{thinking}}\\
    o4-mini-2025-04-16          & \cross   & 76.19 & 23.00 & 55.82 & 31.46 & 38.90 \\
    o3-2025-04-16 & \cross   & \textbf{87.83} & \textbf{31.26} & \underline{60.27} & \textbf{38.77} & \underline{45.94}\\
    Gemini-2.0-Flash(thinking)  & \cross   & 81.22 & 21.13 & 53.72 & 28.84 & 37.26\\
    Gemini-2.5-Flash  & \cross   & 81.48 & 23.74 & 54.72 & 30.71 & 39.05\\
    Gemini-2.5-Pro & \cross   & \underline{83.60} & \underline{29.86} & \textbf{64.47} & \textbf{38.77} & \textbf{46.47}\\
    \midrule

    \rowcolor{lightgray}
    \multicolumn{7}{l}{\textbf{3D visual grounding models: \textit{point clouds, video frames}}}\\
    PQ3D          & \tick    & 30.69 &  \textbf{8.36} & 24.42 & 16.73 & 16.68 \\
    3D-VisTA      & \tick    & 29.10&  \underline{7.44} & 25.05 & 15.98 & 16.26 \\
    Chat-Scene*   & \tick    & \underline{49.10} &  6.55 & \textbf{31.73} & \textbf{22.99} & \textbf{22.90}\\
    Grounded 3D-LLM*   & \tick    & \textbf{49.25} &  6.62 & \underline{26.36} & \underline{19.37} & \underline{20.10}\\
    \midrule
    \rowcolor[gray]{0.7}
    Human** 
                  & –        & 100.00   &  92.00   &  98.00   &  97.00   & 95.00    \\
    \bottomrule
    \end{tabular}

}
\label{table: main results on Anywhere3D-bench}

\end{table}

    \cref{table: main results on Anywhere3D-bench} presents the overall results on our benchmark. Human performance substantially surpasses that of the best-performing models Gemini-2.5-pro and o3, particularly at the space level, indicating that current models fall far short of human-level 3D spatial intelligence.

\paragraph{Area v.s. Space v.s. Object v.s. Part} 
  Grounding expressions at the space level is the most challenging task on our benchmark. This difficulty arises from the need to understand spatial relations and distance, situations, and reason over the absolute locations in 3D space beyond objects. The best-performing model at space level, o3 with visual thinking ability, only achieves 30.73\% accuracy. Part-level visual grounding, though derived from object-level grounding, also poses significant challenges for all models. It requires the model to first identify the object to which the part belongs, and then reason about the part’s location and size based on spatial relationships, functions, and other contextual cues. In contrast, area-level and object-level grounding are relatively easier.

\paragraph{Thinking v.s. Non-thinking}     
    It is worth noting that, in both LLM and MLLM settings, models equipped with thinking capabilities consistently outperform their non-thinking counterparts. For example, Qwen3-32B (thinking) and DeepSeek-R1-671B outperform Qwen3-32B (non-thinking) and DeepSeek-V3-671B across all four grounding levels. Similarly, Gemini-2.0-Flash with thinking enabled surpasses its non-thinking variant by a substantial margin of nearly 9\%. These results highlight the critical role of thinking abilities in effectively tackling our benchmark.

\paragraph{LLMs v.s. MLLMs}
     Additional visual inputs, i.e., video frames and the bird’s-eye view image, consistently improve the performance of the same models (GPT-4.1, o4-mini, and Qwen2.5-VL-72B) when transitioning from the LLM setting to the MLLM setting. The performance gains are notable at object level (8.19\% on average) and part level (4.76\% on average), as models can leverage visual inputs to access richer information about the object's details, such as color and structure. However, improvements at the space level are less pronounced (3.47\% on average), suggesting that current MLLMs have limited ability to interpret spatial relationships in 3D space from 2D images.


\paragraph{MLLMs v.s. 3D Visual Grounding Models}
   Both MLLMs and 3D visual grounding models are provided with visual inputs, along with the ground-truth information of object locations and sizes. However, specialized 3D visual grounding models demonstrate limited performance, particularly at the space level, as they are restricted to predicting objects in the scene and lack generalizability for multi-level visual grounding tasks.

\subsection{Detailed Analysis on Grounding Levels}

Furthermore, we examine model performance across different types of referring expressions at each visual grounding level. Due to space limitations, we leave the analysis on area-level visual grounding to the Appendix \cref{appendix B: detailed analysis on area level}.



\begin{figure}
  \centering
  \begin{minipage}{0.49\linewidth}
    \centering
    \includegraphics[width=\linewidth]{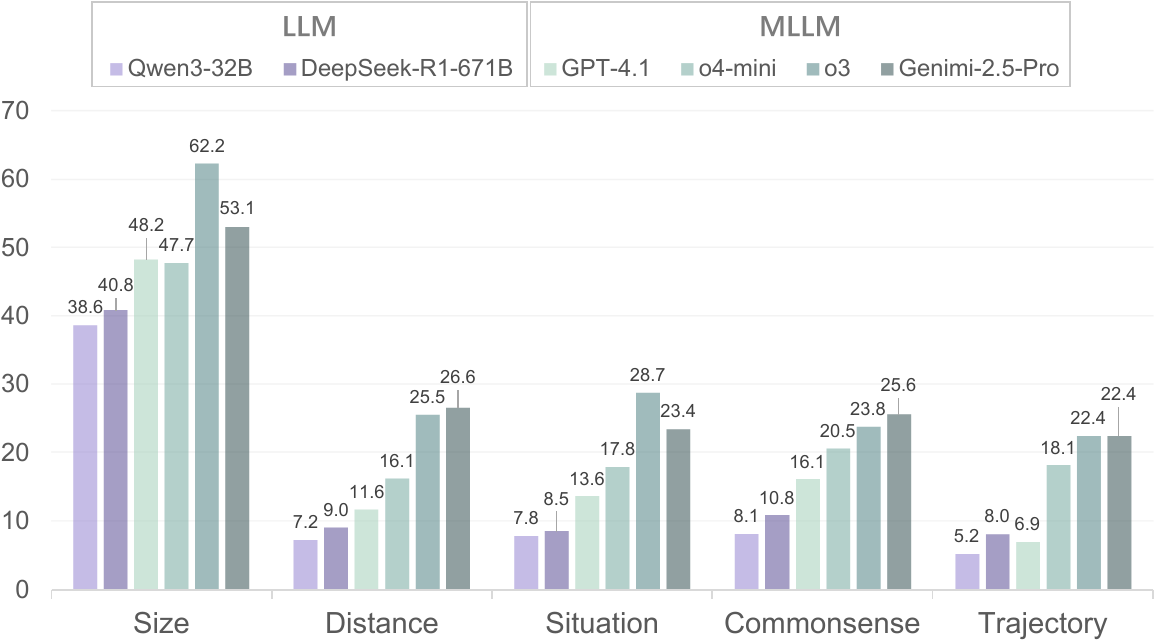}
    \caption{Results on different types of expressions on Space Level.}
    \label{fig:space level detailed analysis}
  \end{minipage}
  \hfill
  \begin{minipage}{0.49\linewidth}
    \centering
    \includegraphics[width=\linewidth]{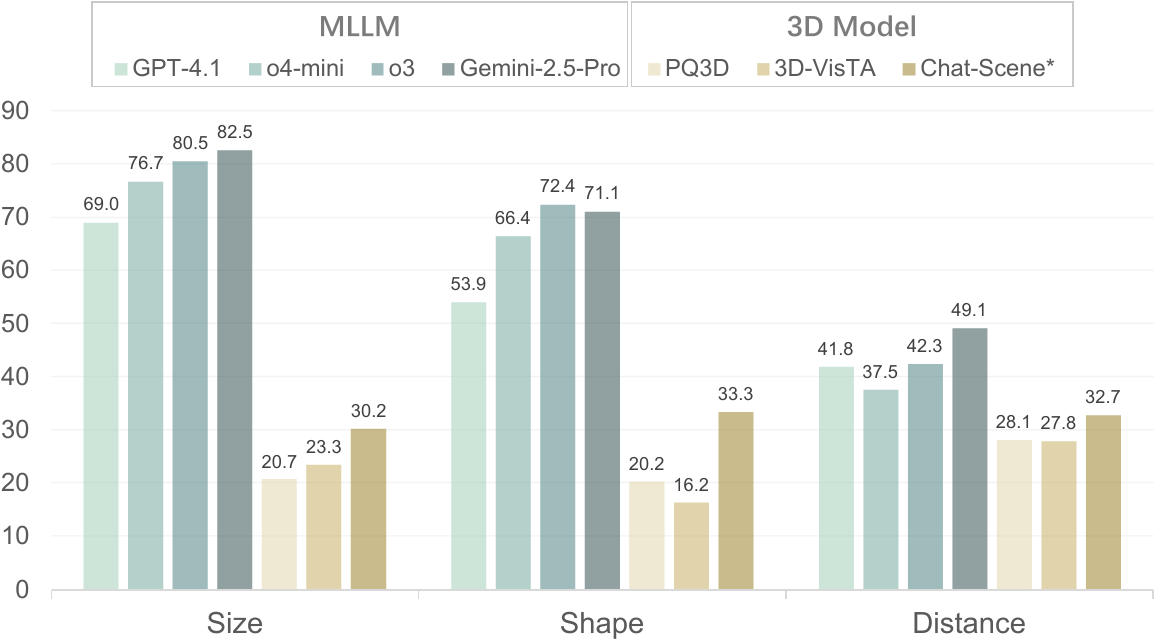}
    \caption{Results on different types of expressions on Object Level.}
    \label{fig:object level detailed analysis}
  \end{minipage}
  \vspace{-1.0em}
\end{figure}

\begin{figure*}[htbp]
  \centering
  \begin{minipage}[t]{0.48\linewidth}
    \vspace{0pt} 
    \centering
    \includegraphics[width=\linewidth]{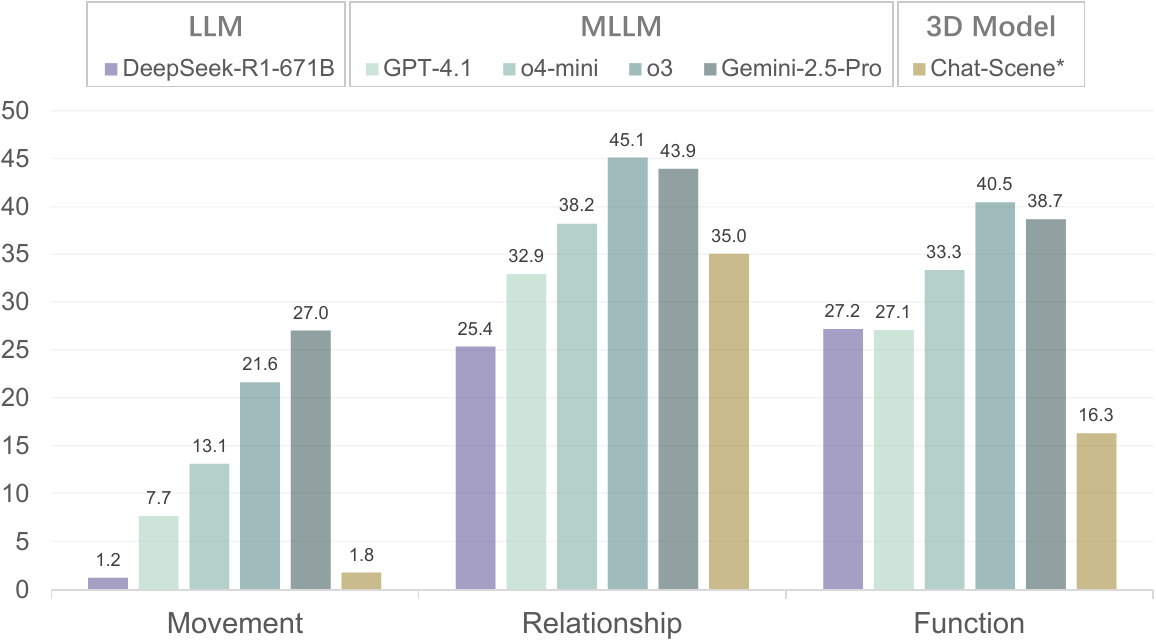}
    \captionof{figure}{Results on different types of expressions on Part Level.}
    \label{fig:part level detailed analysis}
  \end{minipage}
  \hfill
  \begin{minipage}[t]{0.50\linewidth}
    \vspace{0pt}
    \centering
    \includegraphics[width=\linewidth]{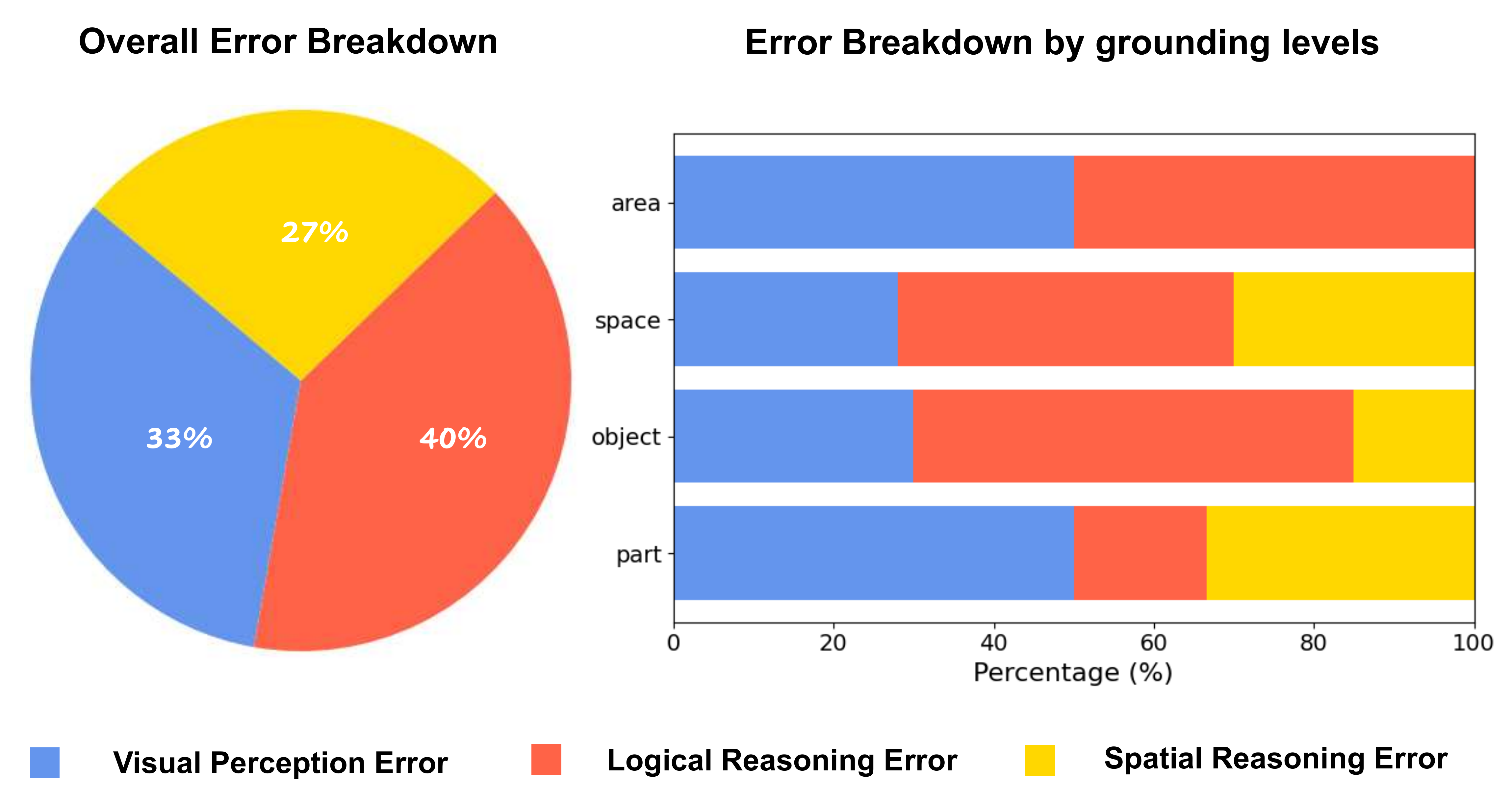}
    \caption{Gemini-2.5-Pro error Breakdown on human-evaluation subset.}
    \label{fig:error breakdown new}
  \end{minipage}
\end{figure*}

\paragraph{Space Level}
    We report the top-performing LLMs and MLLMs on different types of expressions at the space level, as illustrated in \cref{fig:space level detailed analysis}. Trajectory-based expressions are comparatively challenging, as they require a comprehensive understanding of spatial distance, relationships, and orientation. Expressions involving situation, distance, and commonsense also present difficulties, as they demand reasoning about spatial regions beyond objects in the scene. In contrast, size-related expressions are relatively easier: selecting the correct object, adjusting its size based on instructions, and performing positional refinements require less complex spatial perception and reasoning.

\paragraph{Object Level}
    \cref{fig:object level detailed analysis} demonstrates the performance of three types of expressions at object level. In overall, MLLMs outperform 3D models by a large margin. However, an interesting observation is that, compared to MLLMs, 3D visual grounding models exhibit a more balanced capability in interpreting object size, shape, and distances between objects. This may be attributed to the fact that 3D models are typically trained with inputs such as point clouds, which inherently encode spatial coordinates. In contrast, for MLLMs, object sizes are explicitly provided in the scene graph, which contributes to their stronger performance on size-related tasks with Gemini-2.5-pro and o3 achieving move than 80\% accuracy, whereas estimating distances between objects requires both computation and commonsense reasoning, resulting in Gemini-2.5-pro and o3 achieving less than 50\% accuracy on distance-related task.

    Moreover, compared to the visual grounding results reported in ScanRefer~\cite{scanrefer}, where 3D visual grounding specialist models achieve around 50\% accuracy, their performance drops substantially on our benchmark. This suggests that reasoning about quantitative object size and inter-object distance remains a significant challenge for current 3D visual grounding approaches.

\paragraph{Part Level}
    \cref{fig:part level detailed analysis} shows the performance of top-performing models on different types of expressions at part level. Expressions involving dynamic movement present the greatest challenge, with the best-performing model (Gemini-2.5-Pro) achieving only 27\% accuracy, as models must not only accurately identify the specific part of the object but also understand the object's orientation to correctly predict the position of the bounding box after the movement. 
    For expressions centered on spatial relationships, the best-performing non-thinking MLLM (GPT-4.1) is beaten by the best-performing 3D visual grounding model (Chat-Scene*), which simply outputs the entire referenced object, highlighting non-thinking MLLM's limited ability in understanding spatial relationships and object orientations.

\subsection{Error Analysis}
    To further quantitatively analyze the failure cases of the best-performing model, i.e., Gemini-2.5-Pro, we conduct an error analysis on the human evaluation set and categorized the errors into three main types, as illustrated in \cref{fig:error breakdown new}.
    \begin{itemize}
        \item \textbf{Visual Perception Error}: Errors arising from the misrecognition of object categories, attributes, or other visual properties based on visual inputs or object captions.
        \item \textbf{Logical Reasoning Error:} Refers to errors stemming from incorrect logical reasoning or contextual inconsistencies, including incomplete comprehension of the referring expression or commonsense reasoning failures not attributable to visual perception.
        \item \textbf{Spatial Reasoning Error:} Errors arising from incorrect identification of object orientation or inaccurate alignment between egocentric descriptions and the global coordinate system (x, y, z). 
    \end{itemize}

    \begin{figure}[ht]
        \centering
        \includegraphics[width=\linewidth]{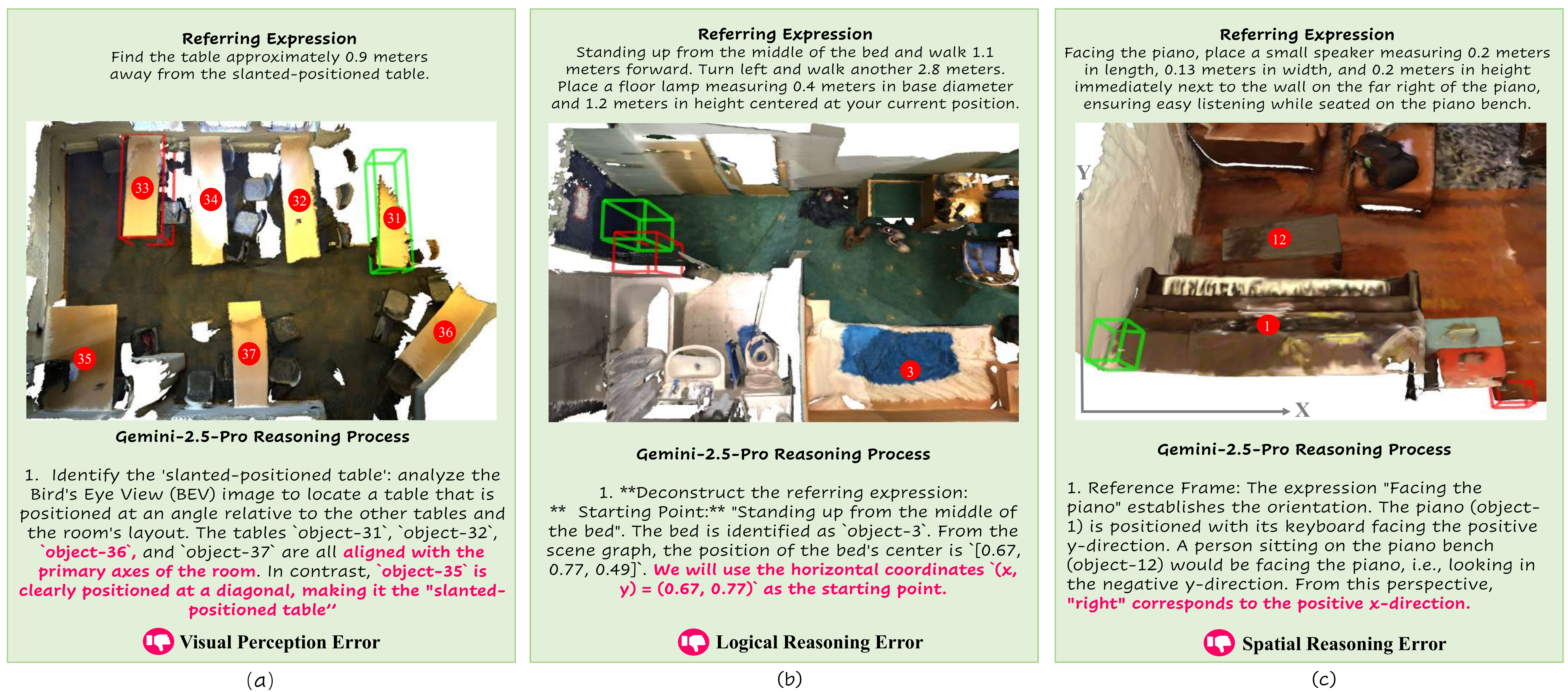}
        \caption{(a): \textcolor{cornflowerblue}{Visual Perception Error}, (b): \textcolor{tomato}{Logical Reasoning Error}, (c): \textcolor{gold}{Spatial Reasoning Error} made by Gemini-2.5-pro on Anywhere3D-Bench. Green bounding boxes represent ground-truth while red boxes represent Gemini-2.5-pro's prediction. The error in reasoning process made by Gemini-2.5-pro is highlighted in \textcolor{pink}{bold}.}
        \label{fig:gemini qualitative results}
    \end{figure}

    Overall, the proportions of the three error types are relatively balanced. However, at the part level, the proportion of perception errors is higher, as this level requires finer-grained visual understanding of object components. In contrast, at the space level, linguistic and spatial reasoning errors are more prevalent, due to the increased demand for spatial orientation and logical inference — particularly in mapping egocentric references (e.g., front, back, left, right) to absolute relationships in the global (x, y, z) coordinate system. We observe that this remains a significant challenge even for Gemini-2.5-pro. 
    
    As illustrated in \cref{fig:gemini qualitative results} (a), Gemini-2.5-Pro fails to visually recognize the slanted table in the bird’s-eye view image. In \cref{fig:gemini qualitative results} (b), it incorrectly treats the starting position of the trajectory as the geometric center of the bed, rather than the middle of the bed’s edge, suggesting that it lacks a human-like understanding of such expressions. In \cref{fig:gemini qualitative results} (c), although the model correctly infers the person’s orientation(``looking in the negative y-direction''), however, it fails to correctly interpret the spatial meaning of ``right'' with respect to this orientation.

\section{Best-performing Thinking Model v.s. Best-performing Non-Thinking Model}

In this section, we further explore where thinking models outperform their non-thinking counterparts on Anywhere3D-Bench.

\subsection{Qualitative Comparisons}
\begin{figure}[ht]
  \centering
    \includegraphics[width=\linewidth]{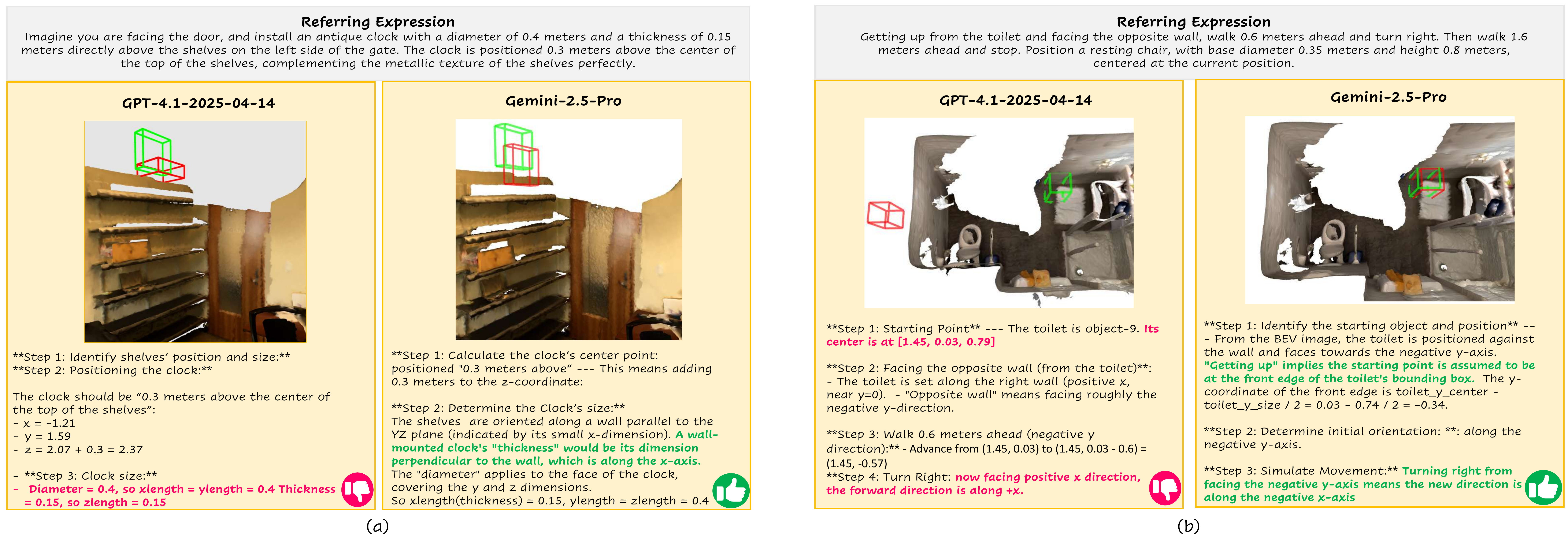}
    \vspace{-1em}
    \caption{Qualitative results on GPT-4.1 and Gemini-2.5-pro reasoning process. Green boxes indicate ground-truth, while red boxes show predictions from GPT-4.1 and Gemini-2.5-pro.}
    \label{fig:Qualitative results fig1 on GPT-4.1 and Gemini 2.5 Pro}
\end{figure}

We begin by presenting some qualitative comparisons between the best-performing non-thinking model (GPT-4.1) and the best-performing thinking model (Gemini-2.5-Pro), as shown in \cref{fig:Qualitative results fig1 on GPT-4.1 and Gemini 2.5 Pro}.

In \cref{fig:Qualitative results fig1 on GPT-4.1 and Gemini 2.5 Pro} (a), the referring expression asks the model to predict the location and size of a newly placed clock above the shelf. While both GPT-4.1 and Gemini-2.5-Pro correctly identify the shelf and estimate the clock’s size, GPT-4.1 misinterprets the clock’s orientation, incorrectly treating its thickness as its height. In contrast, Gemini-2.5-Pro accurately understands the clock’s orientation by reasoning that ``the thickness would be its dimension perpendicular to the wall'' demonstrating stronger spatial reasoning and alignment with the global coordinate axes.

In \cref{fig:Qualitative results fig1 on GPT-4.1 and Gemini 2.5 Pro} (b), which involves trajectory understanding, Gemini-2.5-Pro exhibits better situational reasoning by noting that ``getting up implies the starting point is at the front edge of the toilet'' whereas GPT-4.1 simplistically assumes the starting point is at the center of the toilet. Additionally, Gemini-2.5-Pro more accurately reasons about spatial relationships and directions, correctly stating that ``turning right means the new direction is along the negative x-axis'' while GPT-4.1 fails to infer the correct post-turn direction.

These qualitative results highlight that thinking models are more effective at leveraging visual information to analyze spatial relationships and situational context, whereas non-thinking models struggle to perform such integrative reasoning. More qualitative results are in Appendix \cref{appendix B: more qualitative results on comparison of GPT-4.1 and Gemini-2.5-pro}.

\subsection{Comparisons on spatial reasoning}

Based on the above analysis, GPT-4.1 makes errors when mapping directions(i.e., left, right) to the correct spatial axes(i.e. positive X, negative X, positive Y, or negative Y) in the 3D scene. We further investigate this question type to assess how non-thinking and thinking models differ in performance. We prompt the model to imagine facing a given spatial axis (i.e., positive X, negative X, positive Y, or negative Y) and evaluate its ability to correctly map the terms ``left'' and ``right'' to the corresponding spatial axes under either a right-handed or left-handed coordinate system, as demonstrated in \cref{tab:GPT_4.1_Gemini_2.5_pro}.

    

Anywhere3D-Bench adopts a right-handed coordinate system, which is commonly used in 3D scene. Under this setting, thinking models perform remarkably well, making few errors and significantly outperforming their non-thinking counterparts. This advantage is also evident in the qualitative results \cref{fig:Qualitative results fig1 on GPT-4.1 and Gemini 2.5 Pro} (b). However, when switched to a left-handed coordinate system, both types of models' performance consistently decline, highlighting the limitations of current MLLMs/LLMs in spatial reasoning. Detailed implementation and evaluation results are provided in the Appendix.

\begin{table}[htbp]
  \centering
  \caption{Comparison of different models' ability to map directional terms(``left'', ``right'') to spatial axes under left-handed or right-handed coordinate system}
  \label{tab:GPT_4.1_Gemini_2.5_pro}
  \resizebox{\linewidth}{!}{
  \begin{tabular}{lcccccc}
  \toprule
  Accuracy  & Qwen2.5-72B & Qwen2.5-72B-VL & GPT-4.1 & Gemini-2.5-pro & o3 & DeepSeek-R1-671B\\ 
  \midrule
    left-handed Coor. & 37.50\% & 41.67\% & 50\% & 54.17\% & 75\% & 58.33\%\\
    Right-handed Coor.& 67.50\% & 50\% & 66.70\% & 87.5\%  & 100\% & 58.33\% \\
    
  \bottomrule
  \end{tabular}
  }
  \vspace{-1.5em}
\end{table}

\section{Related Work}

\paragraph{3D Visual Grounding} 3D vision-language learning establishes critical connections between natural language and 3D environments, enabling applications in augmented/virtual reality~\citep{d3net, zhang2024towards} and embodied AI systems~\citep{embodied-survey}. 3D visual grounding---the precise localization of language-referred entities in 3D scenes---has emerged as a cornerstone for spatial intelligence. Despite the proliferation of benchmarks~\citep{scanrefer, referit3d, multi3drefer, arkitscenerefer, scanreason, huang2025unveiling}, existing datasets remain predominantly object-centric, constraining models to coarse-grained scene understanding. Recent efforts like SceneFun3D~\citep{scenefun3d} partially address this limitation by introducing a predefined set constrained on small functional elements (e.g., handles, buttons). In contrast, Anywhere3D involves more open-ended object parts (e.g., toilet tank, lampshade of the lamp) and emphasizes the visual grounding of part movements, as shown in \cref{fig:teaser}. MMScan~\cite{mmscan} introduces region-level visual grounding, which extends object-centric tasks to human-activities regions, similar to our area-level tasks. However, it does not involve visual grounding at \textbf{unoccupied space}, such as placing a new object or moving an existing object to a specified unoccupied space within the scene. Concurrently, while advanced visual grounding methods~\citep{3dvista, pq3d, unit3d, viewrefer, eda, 3dsps, butd, 3dvg, vil3dref, yang2024exploiting, shi2024aware, xu2024multi, lu2024scaneru, zhang2024cross, chang2024mikasa, llm-grounder, yuan2024visual, zhu2025mtu} demonstrate progress in object-level localization, their capacity to interpret referrals at multi-levels remains underexplored. Our benchmark bridges this gap by introducing multi-granular localization across four hierarchical levels---\textit{ area, space, object}, and \textit{part}---systematically evaluating model performance in complex, real-world 3D scene grounding.

\paragraph{Evaluating MLLMs on 3D Spatial Understanding} Recent advancements in LLMs have facilitated their integration into 3D domains. Early approaches, often termed ``3D LLMs,'' such as 3D-LLM, LEO, and Chat-Scene~\citep{3dllm,pointllm,3DMIT,scene-llm,multiply,ll3da,leo,chatscene}, fine-tune LLMs to process embedded 3D object features. However, fine-tuning for 3D tasks is computationally expensive and risks catastrophic forgetting~\citep{zhai2023investigating}. In contrast, GPT4Scene~\citep{gpt4scene} demonstrates that MLLMs can effectively tackle 3D understanding through simple visual prompting, bypassing the need for task-specific adaptation, which highlights the untapped potential of MLLMs in 3D intelligence. Concurrently, there is a growing interest in benchmarking off-the-shelf MLLMs on 3D tasks. VSI-Bench~\citep{thinkinginspace} evaluates 3D spatial reasoning in video understanding, while All-Angles Bench~\citep{all-angles} tests MLLMs' ability to establish correspondence between multi-view visual data. ScanReQA~\citep{ScanReQA} further investigates how multimodal inputs affect spatial reasoning, comparing traditional 3D LLMs and MLLMs. Space3D-Bench~\cite{space3d-bench} encompasses a variety of spatial tasks—including object localization, spatial measurements, and navigation—that span both objects and entire rooms. Despite these efforts, the field has yet to systematically assess the 3D visual grounding abilities of MLLMs, leaving open questions about their precision in localizing and reasoning within complex spatial scenes. For detailed discussion and comparison of these works, please refer to Appendix \cref{appendix: detailed related work}.

\section{Conclusion}
\label{sec:conclusion}

In this paper, we present \textit{Anywhere3D-Bench}, a novel and challenging benchmark that extends visual grounding to four levels in 3D scenes. Evaluation results show that even the best-performing MLLMs, Gemini-2.5-pro and o3 with reasoning capabilities, struggle with the two most difficult tasks: space-level and part-level grounding. This highlights the difficulty current MLLMs face in understanding and reasoning about 3D scenes based on 2D visual inputs. Furthermore, specialized 3D visual grounding models consistently underperform compared to MLLMs, particularly on space-level tasks, revealing their limited generalizability to multi-level grounding tasks.

\newpage

\setcounter{footnote}{0}

\appendix

The Appendix is organized into five sections, following the same structure as the main paper:  
\textbf{Anywhere3D-Benchmark} (Section~\ref{appendix: anywhere3d benchmark}),  
\textbf{Experiments and Results} (Section~\ref{appendix: experiments and results}), \textbf{Detailed Discussion on Related Work} (Section~\ref{appendix: detailed related work}), and  
\textbf{Limitations and Future Work} (Section~\ref{appendix: limitations and future directions}).

\section{Anywhere3D Benchmark}
\label{appendix: anywhere3d benchmark}

\renewcommand\thefigure{A\arabic{figure}}
\setcounter{figure}{0}
\renewcommand\thetable{A\arabic{table}}
\setcounter{table}{0}
\renewcommand\theequation{A\arabic{equation}}
\setcounter{equation}{0}

\subsection{Data Statistics}
\label{appendix A: data statistics}

\begin{figure*}[htbp]
      \centering
      \begin{minipage}[t]{0.48\linewidth}
        \vspace{0pt}
        \centering
        \captionof{table}{Number of referring expressions per grounding level across ScanNet, Multiscan, 3RScan and ARKitScenes.}
        \label{tab:expression number across 4 levels and datasets}
        \renewcommand{\arraystretch}{1.1}
        \resizebox{\linewidth}{!}{

            

            
            \begin{tabular}{lcccc}
            \toprule
            Dataset & Area Level & Space Level & Object Level & Part Level \\
            \midrule
            ScanNet & 93 & 498 & 643 & 245 \\
            MultiScan & 5 & 56 & 17 & 20 \\
            3RScan & 16 & 197 & 92 & 67 \\
            ARKitScenes & 75 & 458 & 202 & 202 \\
            \bottomrule
            \end{tabular}
        }
      \end{minipage}
      \hfill
      \begin{minipage}[t]{0.48\linewidth}
        \vspace{0pt} 
        \centering
        \includegraphics[width=\linewidth]{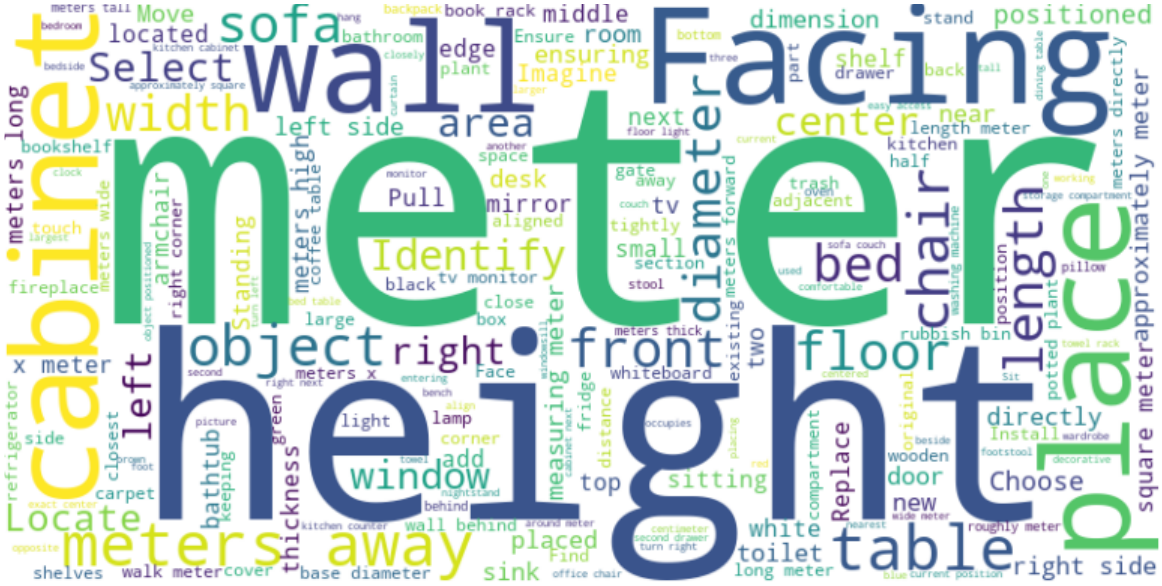}
        \captionof{figure}{Word cloud of \textit{Anywhere3D-Bench}}
        \label{fig:wordcloud}
      \end{minipage}
      
    \end{figure*}

    We first present the number of referring expressions across the four grounding levels on ScanNet, MultiScan, 3RScan, and ARKitScenes, as shown in \cref{tab:expression number across 4 levels and datasets}. To visually demonstrate the linguistic diversity of referring expressions in \textit{Anywhere3D-Bench}, we generate a word cloud based on all expressions, as illustrated in \cref{fig:wordcloud}.

\begin{figure*}[htbp]
      \centering
      \begin{minipage}[t]{0.48\linewidth}
        \vspace{0pt} 
        \centering
        \includegraphics[width=\linewidth]{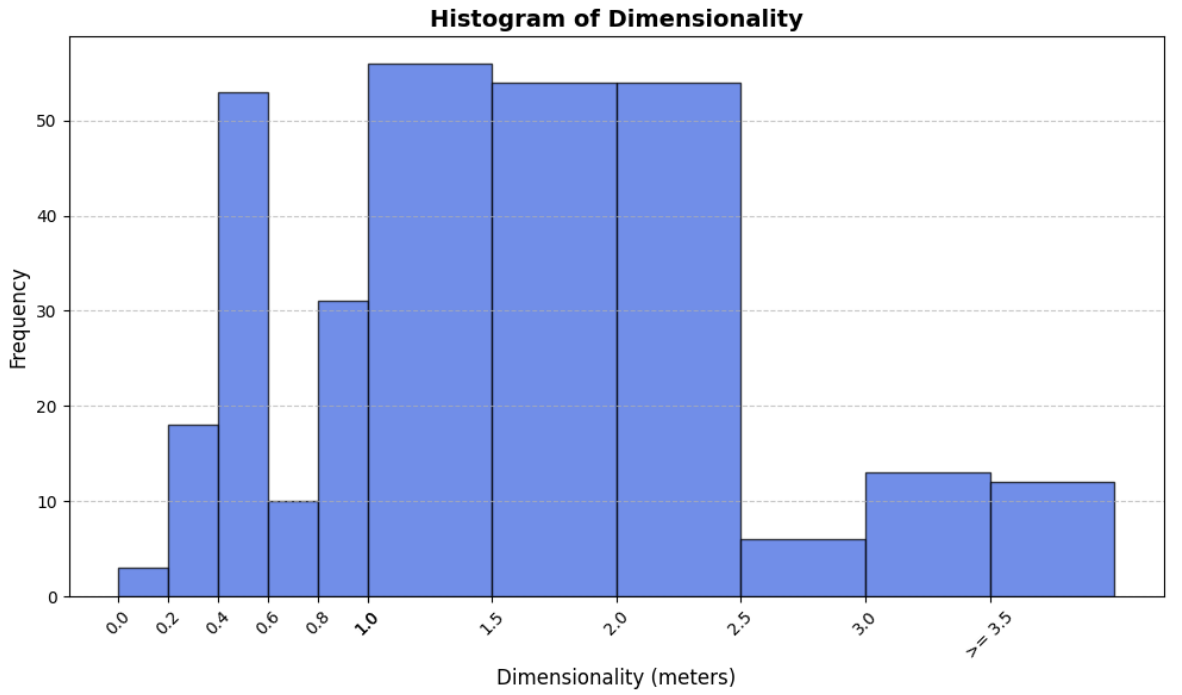}
        \captionof{figure}{Dimensionality Distribution of referring expressions at object level.}
        \label{fig:object level dimensionality distribution}
      \end{minipage}
      \hfill
      \begin{minipage}[t]{0.48\linewidth}
        \vspace{0pt} 
        \centering
        \includegraphics[width=\linewidth]{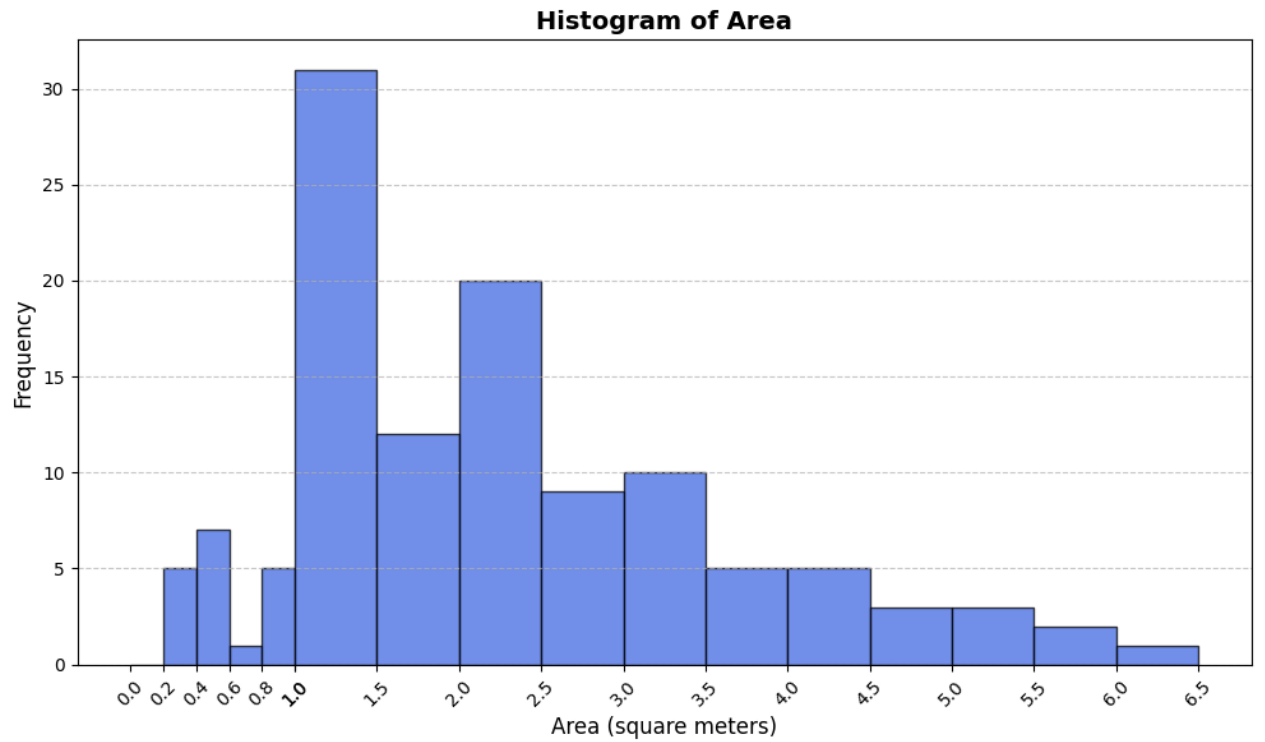}
        \captionof{figure}{Floor Area Distribution of referring expressions at object level }
        \label{fig:object level floor area distribution}
      \end{minipage}
      
    \end{figure*}

\begin{figure*}[htbp]
      \centering
      \begin{minipage}[t]{0.58\linewidth}
        \vspace{0pt} 
        \centering
        \includegraphics[width=\linewidth]{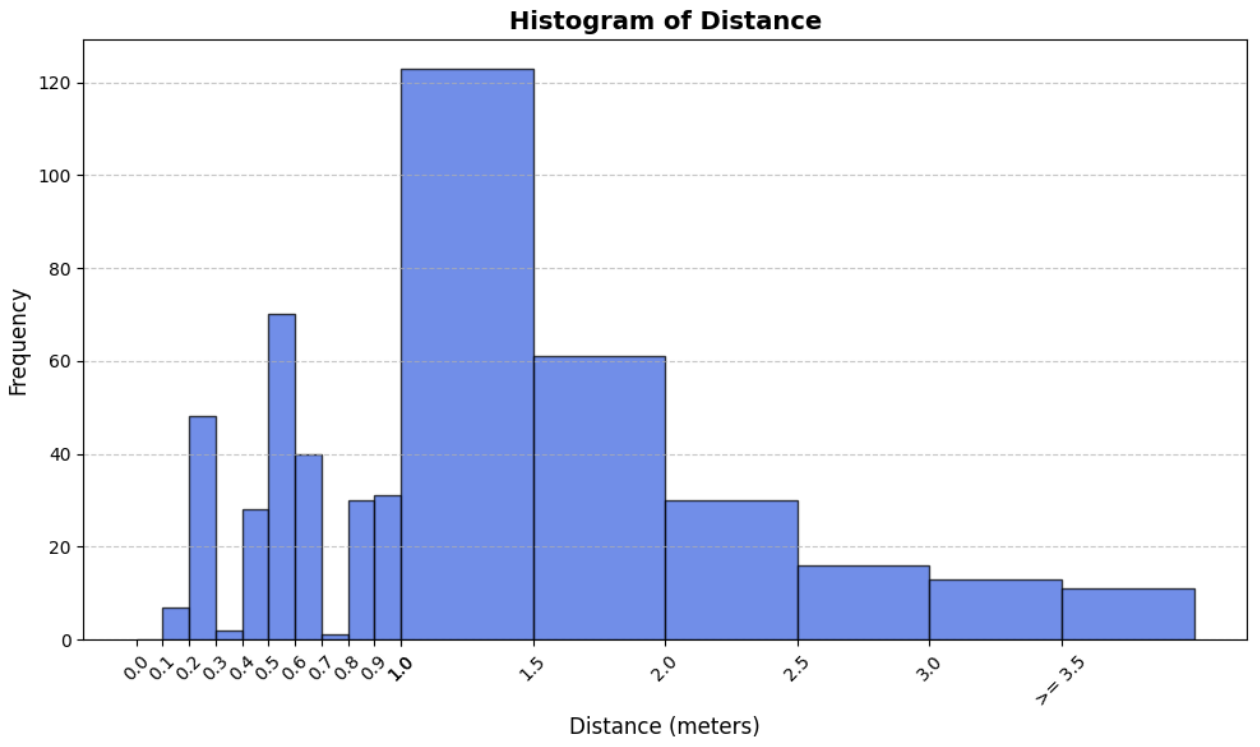}
        \captionof{figure}{Distance Distribution of referring expressions at object level.}
        \label{fig:object level distance distribution}
      \end{minipage}
      \hfill
      \begin{minipage}[t]{0.4\linewidth}
        \vspace{0pt} 
        \centering
        \includegraphics[width=\linewidth]{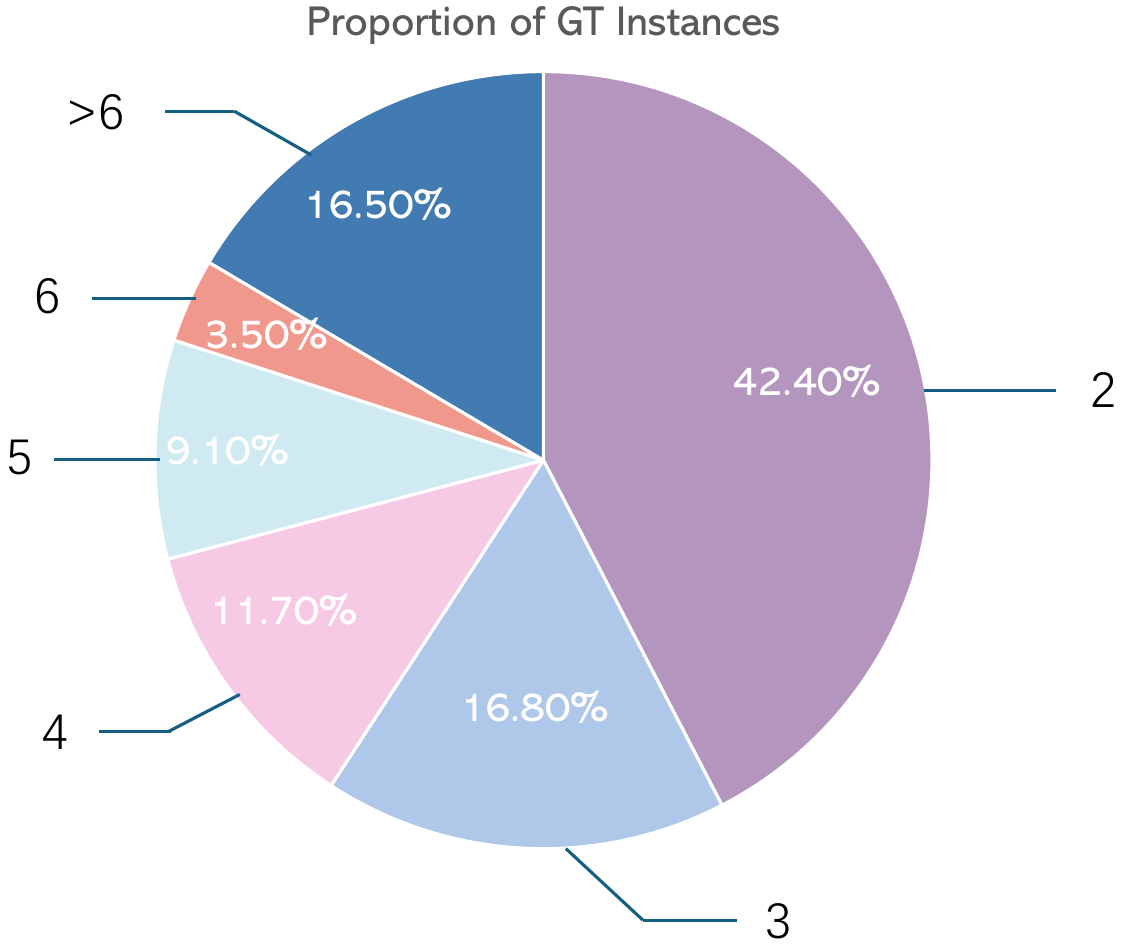}
        \captionof{figure}{Distribution of object categories with two or more ground-truth instances at the object level. Only in these cases is the object label contained in the referring expression.}
        \label{fig:GT instances at object level}
      \end{minipage}
      
    \end{figure*}

   Furthermore, we conduct a distributional analysis of object-level expressions with respect to object size, floor area, and inter-object distance. The results reveal a broad spectrum of referents, ranging from small to large objects and from proximate to distant spatial references, underscoring the diversity of expressions captured in our benchmark (see \cref{fig:object level dimensionality distribution}, \cref{fig:object level floor area distribution}, and \cref{fig:object level distance distribution}).

    To ensure that models predict answer based on understanding and reasoning on object size and inter-object distance, rather than simply matching object labels in the referring expressions, we explicitly exclude ground-truth object labels from the expressions at the object level, unless there are multiple instances of the target object in the scene. This design encourages models to identify the correct object based on 3D scene understanding rather than relying on linguistic cues tied to object categories. As a result, 50\% of the referring expressions do not contain the corresponding ground-truth object label, as they refer to objects with unique instance in the scene. The remaining expressions refer to objects with multiple instances, as illustrated in the distribution shown in \cref{fig:GT instances at object level}.



\begin{figure}
    \centering
    \includegraphics[width=\linewidth]{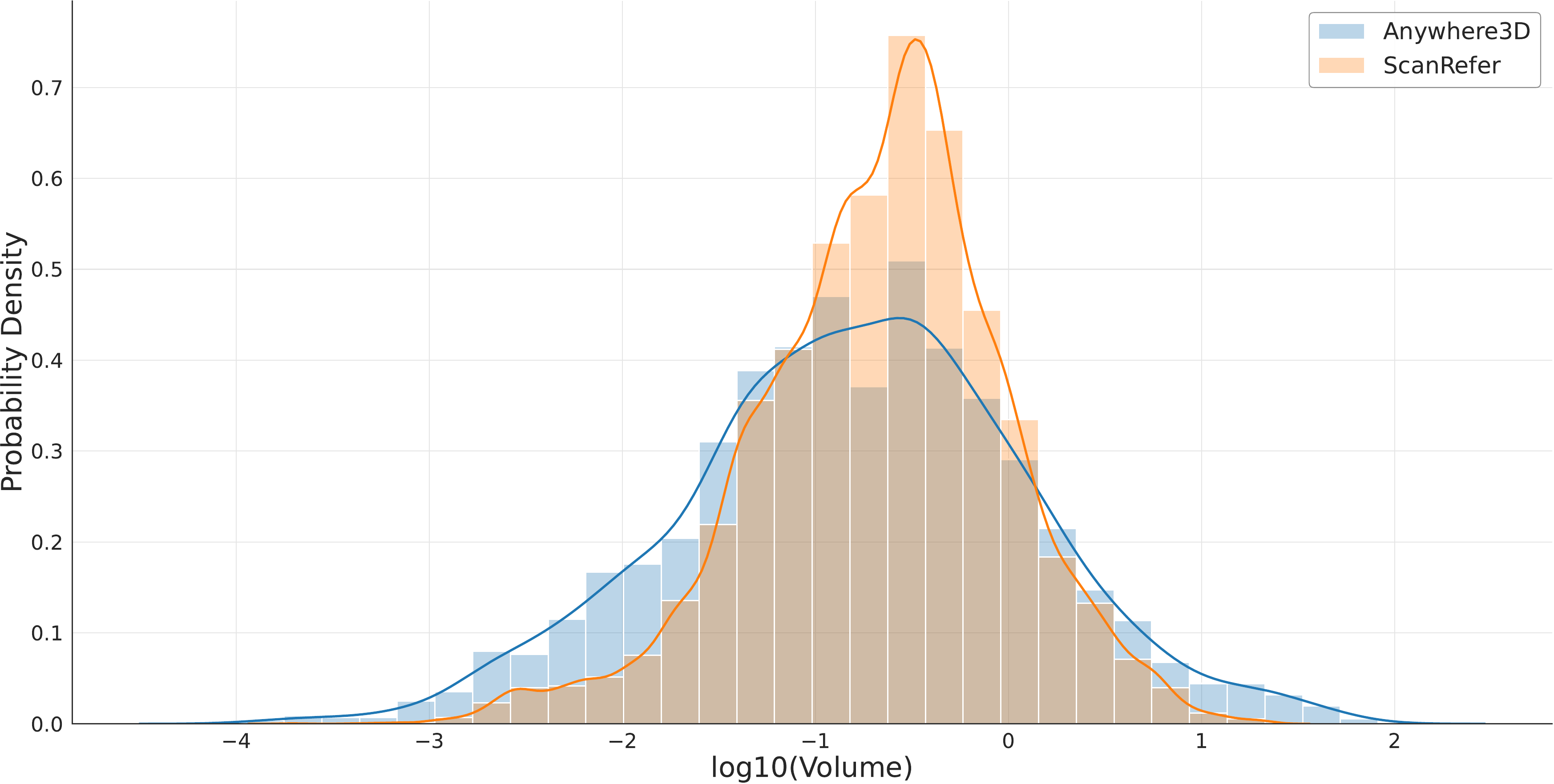}
    \caption{The target objects' volume distribution of Anywhere3D and ScanRefer. Logarithmic scaling is applied to the horizontal axis.}
    \label{fig:volume distribution}
\end{figure}

As shown in \cref{fig:volume distribution}, the volume distribution of Anywhere3D targets exhibits greater dispersion and heavier tails on both ends, indicating more frequent extreme small and large volumes compared to ScanRefer, where volumes are tightly concentrated around the mean. This difference arises because Anywhere3D expands grounding granularity beyond object-level annotations to include spaces, \textit{parts} (typically small), and \textit{areas} (typically large), enabling a more comprehensive evaluation of 3D visual grounding.

\subsection{Data Generation Details}
\label{appendix A: data generation details}

As demonstrated in \cref{fig:prompt messages for referring expressions generation}, \cref{fig:system prompt for expressions generation}, and \cref{fig:few-shot examples for expressions generation}, we present the prompt messages for \textbf{distance-related} referring expressions generation at \textbf{space level} as an example.

\lstdefinestyle{json}{
  basicstyle=\ttfamily\small,
  breaklines=true,
  breakatwhitespace=false,
  showstringspaces=false,
  columns=fullflexible,
  keepspaces=true
}

\begin{figure}[htbp]
\centering
\begin{minipage}{\linewidth}\vspace{0mm}    \centering
\begin{tcolorbox}
\fontsize{7.0pt}{0.8\baselineskip}\selectfont
messages = [\{`role': `system', `content': {\blueprompt{System prompt}}\}, \{`role': `user', `content': \blueprompt{Scene graph of the scene to process}\}]
\end{tcolorbox}
\caption{Prompt messages for referring expression generation with GPT-4o}
\label{fig:prompt messages for referring expressions generation}
\end{minipage}
\end{figure}

\begin{figure}[htbp]
\centering
\begin{minipage}{\linewidth}\vspace{0mm}    \centering
\begin{tcolorbox}
\fontsize{7.0pt}{0.8\baselineskip}\selectfont
You are now a helpful assistant that can generate diverse referring expressions that can be grounded to reasonable space in an indoor scene.\\

The scene is represented by a scene graph in JSON dictionary format. Each entity in the scene graph denotes an object instance, named `<category>-<ID>'. The `caption' field describes the object's attributes, such as `color', `material', etc. The `relations' field specifies the object's spatial relationships with other objects, defined from a viewpoint along the y-axis from positive to negative direction. The `position’ field contains the x, y, z coordinates of object’s center in the scene. The `size’ field describes the width, length and height of the object’s 3D bounding box. The numerical values of `position’ and `size’ correspond to units in meters. For example, from the scene graph:\\

'''

``object\_info'': {``kitchen counter-1'': {``relations'': [``below cabinet-4'', ``lower than soap dispenser-14'', ``lower than paper towel dispenser-15''], ``position'': [1.27, 0.67, 0.9], ``size'': [0.74, 1.86, 0.26], ``caption'': ``The kitchen counter is black granite with a stainless steel sink and faucet ... It is durable, easy to clean, and has a modern, sleek design that matches the stainless steel appliances in the kitchen.''}, ...
}

'''\\

You can know that the center of ``kitchen cabinets-5'' is located in the {x: 1.01, y: 0.37, z: 0.45}, the ``floor-6'' has the width and length of 4.81 meters and 3.05 meters, the ``microwave-8'' is placed within the area of the ``cabinets-3'', the ``cabinet-2'' is to the left of ``water cooler-7'' viewing from +y axis to -y axis, the ``water cooler-7'' is 3 o’clock direction near the ``cabinet-2'' viewing from +y axis to -y axis.\\

Using the provided scene graph, design referring expressions that can be grounded to reasonable space in the 3D scene. There are two principles you need to read and follow very carefully:\\

1. Clarity: Each Referring Expression must be grounded exactly to one target 3D bounding box in space. Do not include the IDs of the objects in the referring expressions. Instead, use ordinal words, colors and relations to refer to different object instances of the same category. Describe the grounding position of the target 3D bounding box using only the surrounding objects to avoid causing confusion. Avoid using terms like ``o' clock'' to describe relations in referring expressions. Do not refer to existing objects with their corresponding positions in the scene! Additionally, consider whether the expressions require a specific viewpoint to ensure the target bounding box is clearly and uniquely identifiable. In some cases, specifying a viewpoint is necessary for achieving this level of precision. Please note that you don't need to stick to the original viewpoint in the scene graph (which is along the y-axis, from the positive to the negative), but if you specify a viewpoint, the spatial relations between objects in your referring expressions need to be consistent with the scene.\\

2. Distance Understanding Related: You should generate referring expressions in which the position of the bounding box should be explicitly specified based on its numerical distance from other objects in the scene. There are two main categories of referring expressions: existing object movement (move object already existed in the scene to another place) and new object increment (Add objects that not exist in the scene). For the category of ‘new object increment’, you may assume objects that not exist in the scene graph in each of the referring expressions. However, they should have reasonable sizes(For example, they should not exceed the boundary of the scene, not overlap with the existing objects.) and should be placed reasonably in the scene (For example, they are not allowed floating in the air). Also, you need to explicitly provide the sizes, i.e. (WIDTH, LENGTH, HEIGHT) or (DIAMETER, HEIGHT). You can also generate referring expressions beyond these categories that meet the principles above.\\

Below are some example referring expressions. Please note that these examples are derived from different scene graphs.\\

\blueprompt{EXAMPLES}\\

After you understand the contents above, I will provide a new scene graph below. Based on the two guiding principles and the examples provided above, generate referring expressions corresponding to the new scene graph.

\end{tcolorbox}
\caption{\blueprompt{System Prompts} for generating distance-related referring expressions at the space Level using GPT-4o}
\label{fig:system prompt for expressions generation}
\end{minipage}
\end{figure}

\begin{figure}[htbp]
\centering
\begin{minipage}{\linewidth}\vspace{0mm}    \centering
\begin{tcolorbox}
\fontsize{7.0pt}{0.8\baselineskip}\selectfont

1. Move the smaller trash bin next to the refrigerator to a spot 0.9 meters directly in front of the bicycle.\\

2. Facing the shower curtain, move the rug on the floor 0.3 meters to the right.\\

3. Pull the piano bench out by 0.5 meters to allow space for someone to sit and play the piano.\\

4. Move the trash can that is closest to the TV to the floor directly in front of the sink, placing it 0.7 meters away from the kitchen cabinet.\\

5. Shift the television on the wall 0.1 meters to the right.\\

6. Facing the tv, move the coffee table 0.55 meters to its left.\\

7. Place a basin with a radius of 0.25 meters and a height of 0.5 meters on the floor, centered 0.35 meters to the right of the sink.\\

8. Place a round coffee table 0.8 meter in front of the couch. The table has a diameter of 1.2 meter and a height of 0.7 meter.\\

9. Move the painting on the wall next to the bed downward so that the bottom of the painting is 1 meter above the floor.\\

10. Add a new table in front of the big sofa with dimensions 80 cm × 60 cm × 1 m. The table should be placed 1.5 meters away from the sofa and centered relative to it.\\

11. Sitting on the chair in front of the desk, place a notebook with 0.4m * 0.5m * 0.1m to the right of the laptop. Keep a distance of 0.15 meters between them.\\

12. Facing the whiteboard, add another foosball table in font of the old one, remaining 0.2 meters between the two tables.\\

\end{tcolorbox}
\caption{\blueprompt{Examples} in \blueprompt{System Prompt} for GPT-4o referring expressions generations in distance-related expressions at space level.}
\label{fig:few-shot examples for expressions generation}
\end{minipage}
\end{figure}

\subsection{Human Annotation and Verification Details}
\label{appendix A: human annotation and verification details}

\subsubsection{Annotation Interface}
\begin{figure}[htbp]
    \centering
    \includegraphics[width=\linewidth]{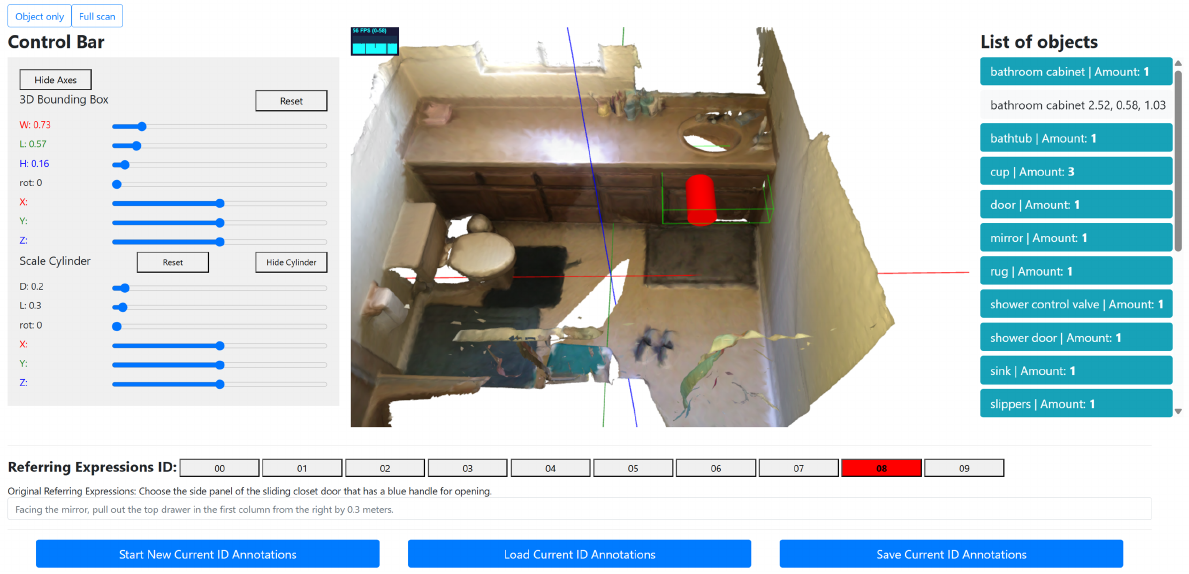}
    \caption{Human Annotation at part level: ``Facing the mirror, pull out the top drawer in the first column from the right by 0.3 meters.''.}
    \label{fig:part level human interface}
\end{figure}

\begin{figure}[htbp]
    \centering
    \includegraphics[width=\linewidth]{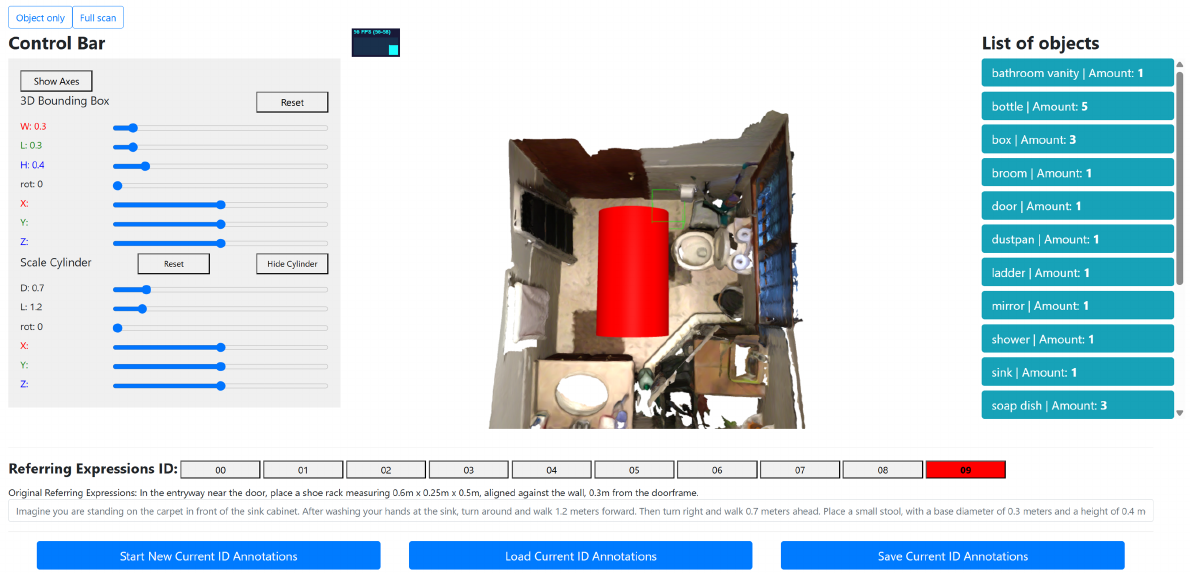}
    \caption{Human Annotation at space level: ``Imagine you are standing on the carpet in front of the sink cabinet. After washing your hands at the sink, turn around and walk 1.2 meters forward. Then turn right and walk 0.7 meters ahead. Place a small stool, with a base diameter of 0.3 meters and a height of 0.4 meters, centered at your current position.'' The scale cylinder serves as good annotation tool for trajectory annotation at space level.}
    \label{fig:space level human interface}
\end{figure}

\cref{fig:part level human interface} and \cref{fig:space level human interface} illustrate the human annotation interface, which is adapted from ScanRefer. Our interface comprises four main components: a control bar, a 3D scene visualization module, an object list, and a referring expression editing box.

The control bar, located in the top-left corner, includes three primary tools: (1) a 3D bounding box annotation tool, (2) a distance measurement tool (i.e. \textit{Scale Cylinder}), and (3) a coordinate axis visualization tool. Both the bounding box and distance measurement cylinder can be resized and repositioned using the control bar, and can also be interactively adjusted with the mouse. The dimensions `W', `L', and `H' represent the lengths of the bounding box along the x-, y-, and z-axes, respectively.

The 3D scene visualization module supports interactive operations such as zooming and rotating, allowing annotators to conduct detailed spatial exploration in the scene.

The object list, located in the top-right corner, displays all objects in the scene, along with their associated labels and sizes.

The referring expression editing box presents the expressions initially generated by GPT-4o and provides an interactive field for manual revision. Annotators can save their annotations or load previously saved ones as needed.

\subsubsection{Statistics on final referring expressions after human verification}

Here, we provide quantitative statistics on the divergence of final expressions after human verification from the original expressions generated by GPT.

Before annotation began, we established clear guidelines: human annotators are allowed to revise the GPT-4o-generated expressions, but they first had to estimate the proportion of modification within the expression. If the required changes exceeded 50\% of the original expression, annotators were allowed to ``skip'' that referring expression. Expressions needing such extensive revision were filtered out.

Overall, 25\% of the candidate expressions were marked as ``skip'' and thus excluded from the dataset. For the remaining expressions, the \textbf{Average Modification Ratio} is approximately 42\%, as calculated by \cref{equ: average modification ratio}:


\begin{equation}
\text{Average Modification Ratio} 
= \frac{1}{N} \sum_{i=1}^{N} 
\frac{\operatorname{Levenshtein}\!\left(E_i^{\mathrm{GPT}}, \, E_i^{\mathrm{Human}}\right)}
{|E_i^{\mathrm{GPT}}|}
\label{equ: average modification ratio}
\end{equation}

where $N$ denotes the total number of referring expressions, $\text{Levenshtein}(E_i^{\mathrm{GPT}}, E_i^{\mathrm{Human}})$ represents the Levenshtein Distance (i.e., the minimum number of single-word edits required to transform the GPT-4o-generated expression $E_i^{\mathrm{GPT}}$ into the final human-verified expression $E_i^{\mathrm{Human}}$), and $|E_i^{\mathrm{GPT}}|$ is the length of the original GPT-4o-generated expression.

\subsubsection{Total Cost and Duration of the Human Annotation \& Verification Process}

Overall, human annotation and verification process for all referring expressions cost approximately 900 USD and took around six weeks, including time for tool familiarization, pilot annotation, formal annotation, and human verification.

\section{Experiments and Results}
\label{appendix: experiments and results}

\renewcommand\thefigure{B\arabic{figure}}
\setcounter{figure}{0}
\renewcommand\thetable{B\arabic{table}}
\setcounter{table}{0}
\renewcommand\theequation{B\arabic{equation}}
\setcounter{equation}{0}

\subsection{Evaluation Metrics}
\label{appendix B: evaluation metrics}
    In this section, we provide a detailed explanation of the $\mathrm{IoU}$ computation formula, as shown in \cref{eq:iou_appendix}.

   \begin{equation}
    \label{eq:iou_appendix}
    \mathrm{IoU} =
    \begin{cases}
    \mathrm{IoU}^{2D}_{xy}, & \text{if } \mathrm{level} = \text{``area''} \\[6pt]
    \mathrm{IoU}^{2D}_{\setminus i} \cdot \mathbf{1}_{\left\{
    |\mathrm{center}_i^{\mathrm{gt}} - \mathrm{center}_i^{\mathrm{pred}}| < t \;\land\;
    \mathrm{size}_i^{\mathrm{pred}} < t
    \right\}}, &
    \parbox[t]{0.45\textwidth}{
    \raggedright
    if $\mathrm{level} \neq \text{``area''}$, \\
    $\mathrm{size}_i^{\mathrm{gt}} < t$, $i \in \{x, y, z\}$
    } \\[6pt]
    \mathrm{IoU}^{3D}, & \text{otherwise}
    \end{cases}
    \end{equation}

\begin{figure}
        \centering
        \begin{minipage}[t]{0.48\linewidth}
            \vspace{0em}
            \centering
            \includegraphics[width=\linewidth]
            {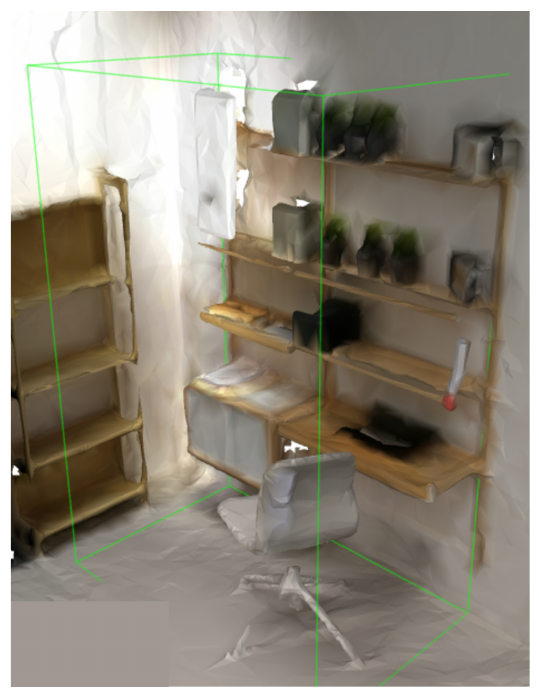}
            \caption{``Choose the area suitable for a study or home office setting.''}
            \label{fig:area level IoU explanation}
        \end{minipage}
        \begin{minipage}[t]{0.48\linewidth}
            \vspace{0.0em}
            \centering
            \includegraphics[width=1\linewidth]{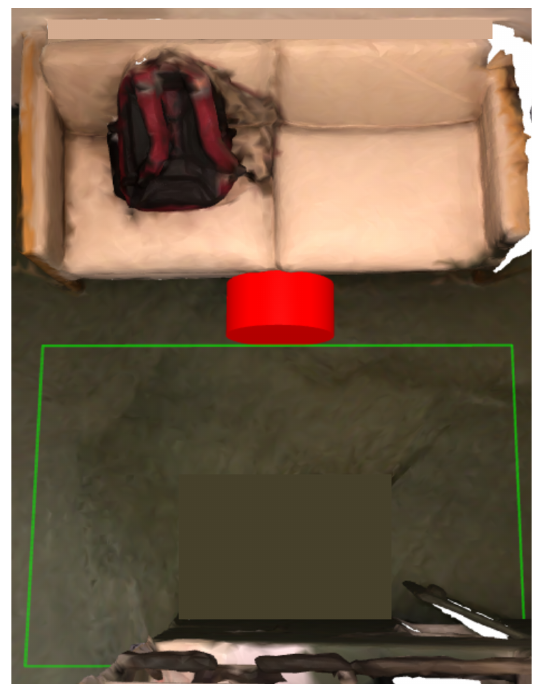}
            \caption{``Lay a new rectangular carpet on the floor directly in front of the sofa, ensuring it is 0.2 meters away from the sofa. The carpet should measure 1.2 meters wide, 1.8 meters long, and \textbf{0.02 meters thick.}'' }
            \label{fig: thin object explanation}
        \end{minipage}

    \end{figure}

    For the area level, we adopt $\mathrm{2DIoU}$ on the horizontal plane as the evaluation metric, due to the inherent ambiguity in defining the vertical extent of the bounding box. For instance, it is unclear whether a study area comprising a desk and an office chair should extend vertically to the ceiling, as illustrated in \cref{fig:area level IoU explanation}.

    Moreover, due to inevitable rendering artifacts in mesh visualization during human annotation, there can be small positional deviations---typically within a few centimeters. For certain objects with very small dimensions along one axis (e.g., a floor carpet that is only 0.02 meters thick(\cref{fig: thin object explanation}), or a wall clock with a thickness of 0.03 meters), directly computing the 3D IoU can lead to significantly distorted results, as the metric becomes overly sensitive to minor positional misalignments. To address this, we design a customized evaluation strategy for such objects.

    Specifically, given a predefined threshold, if the ground-truth bounding box has a dimension smaller than this threshold along any axis (x, y, or z), and if the predicted bounding box satisfy both following conditions: (1) deviates from the ground-truth bounding box by less than the threshold along that axis in terms of position, and (2) is also smaller than the threshold in size along that axis,
    we consider the prediction to be accurate along that axis in both position and size. In this case, the IoU is computed as a 2D IoU over the remaining two axes only. We set the value of threshold to 0.05 meters in the paper.

    \cref{tab: IoU example} illustrates the ground-truth 3D bounding box and predicted 3D bounding box corresponding to \cref{fig: thin object explanation}. The height of the carpet(i.e., the size in the z-axis) is smaller than 0.05 meters. The predicted bounding box satisfies both of the conditions above, so the IoU between the predicted bounding box and the ground-truth bounding box will be the 2DIoU in the XY-Plane. 

    \begin{table}[htbp]
    \centering
    \caption{3D bounding box of ground-truth and prediction of GPT-4.1 corresponding to \cref{fig: thin object explanation}. The predicted 3D bounding box meets the criteria mentioned above, so IoU between ground-truth and prediction will be 2D IoU on the x-y plane $\approx \mathbf{0.4696}$}
    \label{tab: IoU example}
    \begin{tabular}{ccccccc}
    \toprule
                  & \textbf{center\_x} & \textbf{center\_y} & \textbf{center\_z} & \textbf{size\_x} & \textbf{size\_y} & \textbf{size\_z} \\ \midrule
    ground-truth & -0.01     & -1.03     & -0.03     & 1.8     & 1.2     & 0.02    \\
    prediction    & -0.01     & -0.68     & 0.01      & 1.2     & 1.8     & 0.02    \\ 
    \bottomrule
    \end{tabular}
\end{table}

\subsection{Baseline Settings}
\label{appendix B: Baseline Settings}

To encourage diverse reasoning, we use the default temperature settings for all LLM and MLLM models. Each model is evaluated independently over three runs and the mean values are reported in the main paper based on these runs. \cref{tab: model's version} illustrates the specific versions of the LLMs and MLLMs used.

\begin{table}[htbp]
    \centering
    \caption{Versions of LLMs and MLLMs.}
    \label{tab: model's version}
    \resizebox{\linewidth}{!}
    {
    \begin{tabular}{@{} l c c c c c c c c @{}}
  \toprule
  \\
  \textbf{Model} 
    & \textbf{Qwen2.5-72B} 
    & \textbf{Qwen3-32B} 
    & \textbf{DeepSeek-R1-671B} 
    & \textbf{DeepSeek-V3-671B}
    & \textbf{GPT-4.1} 
    & \textbf{o4-mini} 
    & \textbf{o3}
    & \textbf{Qwen2.5-VL-72B} \\
  \midrule
  Version             & 2024-09-19 & 2025-04-29 & 2025-01-20 & 2024-12-26 & 2025-04-14 & 2025-04-16 & 2025-04-16 & 2025-01-27 \\
  
  \bottomrule
\end{tabular}
    }
    
\end{table}

For both LLMs and MLLMs, we design system prompts tailored to each visual grounding level. The object captions in scene graph are generated by Qwen2.5-VL-72B (as further illustrated in Appendix \cref{appendix B: Object Caption Generation by qwen2.5-vl}), one of the strongest open-source vision-language models available at the time. Our motivation for using Qwen2.5-VL-72B as the captioner lies in its open-source availability and to ensure reproducibility. In addition to textual inputs, MLLMs also receive a Bird’s Eye View (BEV) image and eight uniformly sampled video frames as visual inputs, following the setup of GPT4Scene. \cref{fig:prompts for LLMs generated 3D bounding box} and \cref{fig:system prompt for LLMs prediction} show the textual inputs for LLMs at \textbf{space level}, while \cref{fig:prompts for MLLMs generated 3D bounding box} and \cref{fig:system prompt for MLLMs prediction} present the textual inputs for MLLMs at \textbf{space level}. An example of the BEV and video frames used as visual inputs for MLLMs is shown in \cref{fig:gpt4scene bev and frames}.

\begin{figure}[htbp]
\centering
\begin{minipage}{\linewidth}\vspace{0mm}    \centering
\begin{tcolorbox}
\fontsize{7.0pt}{0.8\baselineskip}\selectfont
messages = [\{`role': `system', `content': {\blueprompt{System prompt for specific grounding level}}\}, \{`role': `user', `content': ``Object\_info'': \blueprompt{Scene graph of the scene} + \texttt{``\textbackslash n''} + ``Referring\_expression'': \blueprompt{referring expression}\}]
\end{tcolorbox}
\caption{textual inputs for LLMs}
\label{fig:prompts for LLMs generated 3D bounding box}
\end{minipage}
\end{figure}

\begin{figure}[htbp]
\centering
\begin{minipage}{\linewidth}\vspace{0mm}    \centering
\begin{tcolorbox}
\fontsize{7.0pt}{0.8\baselineskip}\selectfont
You are now a helpful assistant capable of grounding referring expressions to specific space within a 3D scene.\\
\\
The scene is represented by a scene graph in JSON dictionary format. Each entity in the scene graph denotes an object instance, named '<object>-<ID>'. The 'position' field contains the x, y, z coordinates of object’s 3D bounding box center in the scene. The 'size' field indicates the length of the object’s 3D bounding box along the x, y, and z axes. Note that the x and y axes correspond to the horizontal plane, while the z axis corresponds to the vertical direction. The numerical values of 'position', 'size' correspond to units in meters. The "caption" field contains textual description of the object generated by a vision-language model (VLM) based on several images of the object. For example, from the scene graph:\\

'''

``object\_info'': {``object-1'': {``position'': [1.27, 0.67, 0.9], ``size'': [0.74, 1.86, 0.26], ``caption'': ``The kitchen counter is black granite with a stainless steel ...''}, ...
}

'''\\

You can know that the center of ``object-5" is located in the {x: 1.01, y: 0.37, z: 0.45}, the ``object-6'' has the length and width of 4.81 meters and 3.05 meters.\\

``Referring expressions'' are natural language descriptions that point to a specific space within a 3D scene, which is represented by a scene graph. \\

For example, a referring expression like ``Facing the bed, move the nightstand 0.3 meters backward.'' requires identifying the 3D bounding box of the nightstand after it has been moved 0.3 meters backward. \\

Your task is to determine the 3D bounding box corresponding to the referring expression and return the following details:\\

1. The x, y, z coordinates of the center of the bounding box.\\

2. The lengths of the bounding box along the x, y, and z axes.\\

After reviewing the information above, I will provide a new scene graph and a referring expression. Your task is to identify the 3D bounding box that corresponds to the referring expression within the new scene graph.\\

At the end of your response, please provide the following details for the identified 3D bounding box:\\

1. The x, y, z coordinates of its center, formatted strictly as: \{xcoordinate: , ycoordinate: , zcoordinate: \}\\

2. The length of the 3D bounding box along the x, y, and z axes, formatted strictly as: \{xlength: , ylength: , zlength: \}
\end{tcolorbox}
\caption{\blueprompt{System Prompts} of LLMs textual inputs for space level}
\label{fig:system prompt for LLMs prediction}
\end{minipage}
\end{figure}

\begin{figure}[htbp]
\centering
\begin{minipage}{\linewidth}\vspace{0mm}    \centering
\begin{tcolorbox}
\fontsize{7.0pt}{0.8\baselineskip}\selectfont
messages = [\{`role': `system', `content': {\blueprompt{System prompt for specific grounding level}}\}, \{`role': `user', `content': ``Object\_info'': \blueprompt{Scene graph of the scene} + \texttt{``\textbackslash n''} + ``Referring\_expression'': \blueprompt{referring expression} + \texttt{``\textbackslash n''} + ``The subsequent images include a \blueprompt{Bird Eye View} image as the first, followed by 8 \blueprompt{frames} extracted from the scene video. Please return the center coordinates and sizes of predicted 3D bounding box STRICTLY following the instructed format.''\}]
\end{tcolorbox}
\caption{textual inputs for MLLMs}
\label{fig:prompts for MLLMs generated 3D bounding box}
\end{minipage}
\end{figure}

\begin{figure}[htbp]
\centering
\begin{minipage}{\linewidth}\vspace{0mm}    \centering
\begin{tcolorbox}
\fontsize{7.0pt}{0.8\baselineskip}\selectfont
You are a helpful assistant skilled in grounding referring expressions to specific space within a 3D scene.\\
\\
Each scene is represented by the following elements:\\
\\
1. scene graph: a JSON-formatted dictionary that enumerates all objects in the scene. Each entity in the scene graph denotes an object instance, named `<object>-<ID>'. For each object, the `position' field contains the x, y, z coordinates of the center of its 3D bounding box. The `size' field indicates the length of the 3D bounding box along the x, y, and z axes. The x and y axes represent the horizontal plane, while the z axis represents the vertical direction. The values in `position' and `size' are in meters. The ``caption'' field contains textual description of the object generated by a vision-language model (VLM) based on several images of the object.\\
\\
2. Bird’s Eye View (BEV) Image: A top-down view of the scene, where objects’ IDs are labeled in the image.\\
\\
3. 2D Images: A set of 8 frames captured at equal intervals from the scene video. Each frame contains several objects with their object IDs labeled within red circles.\\
\\
Note that object IDs are consistent across the scenegraph, 2D images, and BEV image.\\
\\
Referring expressions are natural language descriptions that point to specific space within the 3D scene. \\
\\
For example, a referring expression like ``Facing the bed, move the nightstand 0.3 meters backward.'' requires identifying the 3D bounding box of the nightstand after it has been moved 0.3 meters backward. \\
\\
Your task is to determine the position and size of the 3D bounding box corresponding to the referring expression.\\
\\
After reviewing the information, I will provide a scene graph, 2D images, and a BEV image of a new scene, along with a referring expression. Your goal is to identify the 3D bounding box that corresponds to the referring expression.\\
\\
At the end of your response, please provide the following details for the identified 3D bounding box:\\
\\
1. The x, y, z coordinates of its center, strictly formatted as: {xcoordinate: , ycoordinate: , zcoordinate: }\\
\\
2. The length of the 3D bounding box along the x, y, and z axes, strictly formatted as: \{xlength: , ylength: , zlength: \}
\end{tcolorbox}
\caption{\blueprompt{System Prompts} of MLLMs textual inputs for space level}
\label{fig:system prompt for MLLMs prediction}
\end{minipage}
\end{figure}

\begin{figure}
    \centering
    \includegraphics[width=0.9\linewidth]{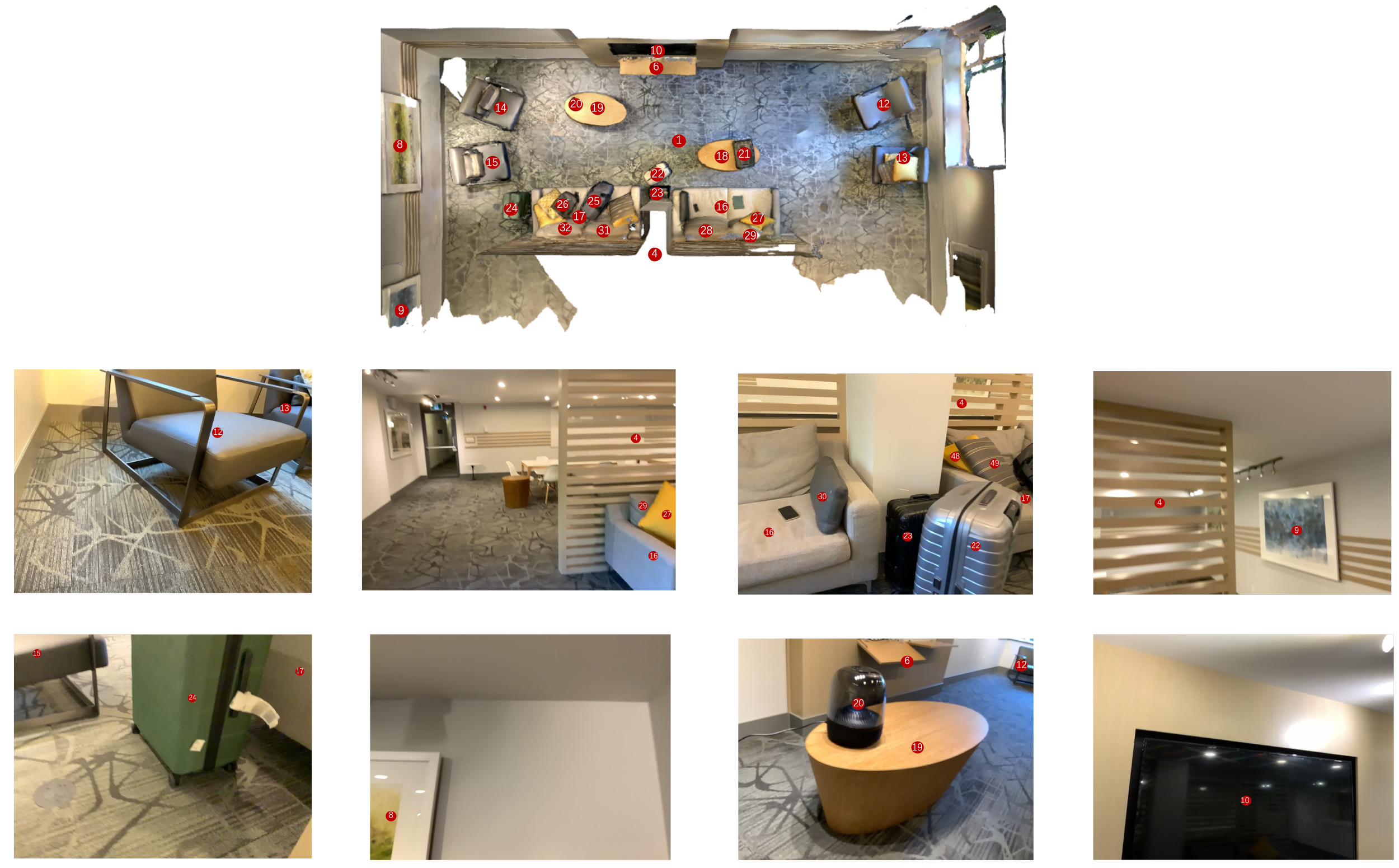}
    \caption{GPT4Scene Bird’s Eye View (BEV) and eight uniformly sampled video frames from MultiScan Scene00109\_00. The image at the top depicts the BEV, while the 2 × 4 grid below shows the video frames. In both the BEV and the video frames, object labels are marked with red circles to indicate object locations, following the visual input in GPT4Scene.}
    \label{fig:gpt4scene bev and frames}
\end{figure}

Notably, we provide both LLMs and MLLMs with the ground-truth locations and sizes of objects. This design decouples object detection from 3D visual grounding, allowing us to specifically examine the models’ capabilities in perceiving, understanding, and reasoning within 3D scenes under ideal localization conditions.

After obtaining the model’s output, we extract the predicted 3D bounding box location and size using a combination of regular expressions and LLM-based parsing.

We utilize the fine-tuned checkpoints released by LLaVA-OneVision, GPT4Scene, PQ3D, 3D-VisTA, Chat-Scene, and Grounded 3D-LLM. Among them, GPT4Scene, PQ3D, 3D-VisTA, Chat-Scene, and Grounded 3D-LLM can only predict object IDs corresponding to the referring expressions. Therefore, we derive the predicted 3D bounding boxes based on the size and location of the corresponding objects in the scene graph. Moreover, Grounded 3D-LLM fails to produce visual grounding results on some scenes of ScanNet due to feature misalignment. As a result, our evaluation is restricted to those scenes where results can be successfully generated.

\subsection{Object Caption Generation for Benchmarking}
\label{appendix B: Object Caption Generation by qwen2.5-vl}
For LLMs and MLLMs, as outlined in \cref{appendix B: Baseline Settings}, the inputs include object captions generated by Qwen2.5-VL-72B. For each object in the scene, we prompt Qwen2.5-VL-72B to generate a descriptive caption using up to \textbf{five} \textit{uniformly} sampled frames where the object is visible. To guide the captioning process, we annotate each frame with a green bounding box—projected from 3D space—to highlight the target object.

The prompt instructs Qwen to describe the object within the bounding box, covering its category, material, color, shape, structure, function, and surrounding environment. The full prompt template is provided in \cref{fig:prompts for Qwen generating object captions}.

\begin{figure}
    \centering
    \includegraphics[width=0.5\linewidth]{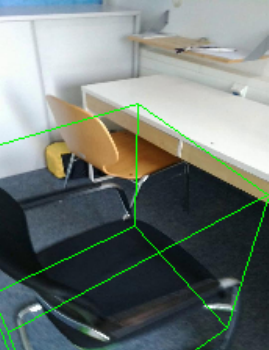}
    \caption{An example of Qwen-generated caption: ``A black plastic chair with a cushioned seat and backrest. The frame is metallic, featuring curved legs and armrests. Designed for ergonomic seating in office or classroom settings. Positioned near a desk with papers and books, suggesting a workspace environment.''}
    \label{fig:qwen caption}
\end{figure}

An example is illustrated in \cref{fig:qwen caption}

\begin{figure}[htbp]
\centering
\begin{minipage}{\linewidth}\vspace{0mm}    \centering
\begin{tcolorbox}
\fontsize{7.0pt}{0.8\baselineskip}\selectfont
Analyze the object in the green bounding box across multiple viewpoints. Generate a caption with:\\
\\
1. **Category**: \\
   - Use specific name if confident (e.g., "mug")\\ 
   - Use "{{generic}}-type object" for uncertain cases (e.g., "container-type object")\\
   - Flag low-confidence predictions with "possibly"\\
\\
2. **Attributes**:\\
   - Color patterns \& material properties\\
   - 3D shape characteristics \& structural features\\
   - Functional affordances\\
   - Contextual placement \& surrounding objects\\
\\
3. **Confidence Signals**:\\
   - "The tapered rim suggests..." \\
   - "While resembling a vase, the presence of..."\\
   - "Inconclusive evidence for..."\\
\\
Output template: \\
"A [material] [confidence][category] with [color] and [shape]. [Structure/texture details]. [Function inference]. [Contextual placement]."\\
\\
Examples:\\
1. High Confidence: "A glossy ceramic mug with deep blue coloring and rounded shape. Has a comfortable handle and flat base. Made for holding hot drinks. Sitting beside a coffee maker and jar of beans on a kitchen counter."\\
\\
2. Moderate Confidence: "A possibly glass vase-type object with translucent amber coloring. Fluted body shape and water droplets on surface. Likely floral display container. Found on windowsill with plants and pruning shears nearby."\\
\\
3. Low Confidence: "A metallic tool-type object, matte silver with angular grooves. Ambiguous function between wrench or specialized clamp. Seen in workshop environment near assembly parts."
\end{tcolorbox}
\caption{Prompts for Qwen generating object captions.}
\label{fig:prompts for Qwen generating object captions}
\end{minipage}
\end{figure}

\subsection{Detailed Analysis on Area Level}
\label{appendix B: detailed analysis on area level}

At the area level, we further categorize referring expressions into two types. The first type is \textit{Objects Combination}, where most of the relevant objects are explicitly mentioned in the expression. For example, \textit{``Identify the conference area with the long rectangular wooden table surrounded by chairs, used for meetings and discussions.''} This expression explicitly refers to the area formed by the table and surrounding chairs. The second type is \textit{Commonsense Reasoning}, where models are required to apply commonsense knowledge to infer implicitly indicated objects before identifying the corresponding area. For instance, \textit{``Choose the conference area suitable for holding face-to-face meetings.''} the expression does not directly mention the objects, and the model must first deduce the components that constitute such a conference area.

\begin{table}[htbp]
    \centering
    \caption{Analysis on area level}
    \label{tab: analysis area level}
    \resizebox{\linewidth}{!}
    {
    \begin{tabular}{@{} l c c c c c c @{}}
  \toprule
  & \multicolumn{3}{c}{\textbf{LLM setting}} 
  & \multicolumn{3}{c}{\textbf{MLLM setting}} \\
  \cmidrule(lr){2-4} \cmidrule(lr){5-7}
  \textbf{Model} 
    & \textbf{Qwen2.5-72B} 
    & \textbf{Qwen3-32B} 
    & \textbf{DeepSeek-R1-671B} 
    & \textbf{GPT-4.1} 
    & \textbf{o4-mini} 
    & \textbf{Qwen2.5-VL-72B} \\
  \midrule
  Object Combination             & 66.15 & 64.62 & 84.62 & \textbf{95.38} & \underline{87.69} & 64.62 \\
  Commonsense Reasoning & 58.06 & 53.23 & 66.94 & \textbf{76.61} & \underline{72.58} & 53.22 \\
  \bottomrule
\end{tabular}
    }
\end{table}

\cref{tab: analysis area level} demonstrates the performance of six models on two types of expressions at area level. All models exhibit weaker performance on expressions that necessitate commonsense reasoning for the initial identification of the target objects, illustrating their limitations in applying commonsense knowledge within 3D scenes.

\subsection{Evaluation Cost}
\label{appendix B: evaluation cost}
The inference cost varies significantly across models, with DeepSeek-R1-671B costing approximately \$40, GPT-4.1 \$53, o4-mini \$120, and Qwen3-32B around \$9 per full evaluation on \textit{Anywhere3D-Bench}.

\subsection{More qualitative results on comparison of GPT-4.1 and Gemini-2.5-Pro}
\label{appendix B: more qualitative results on comparison of GPT-4.1 and Gemini-2.5-pro}

\begin{figure*}[htbp]
  \centering
    \includegraphics[width=\linewidth]{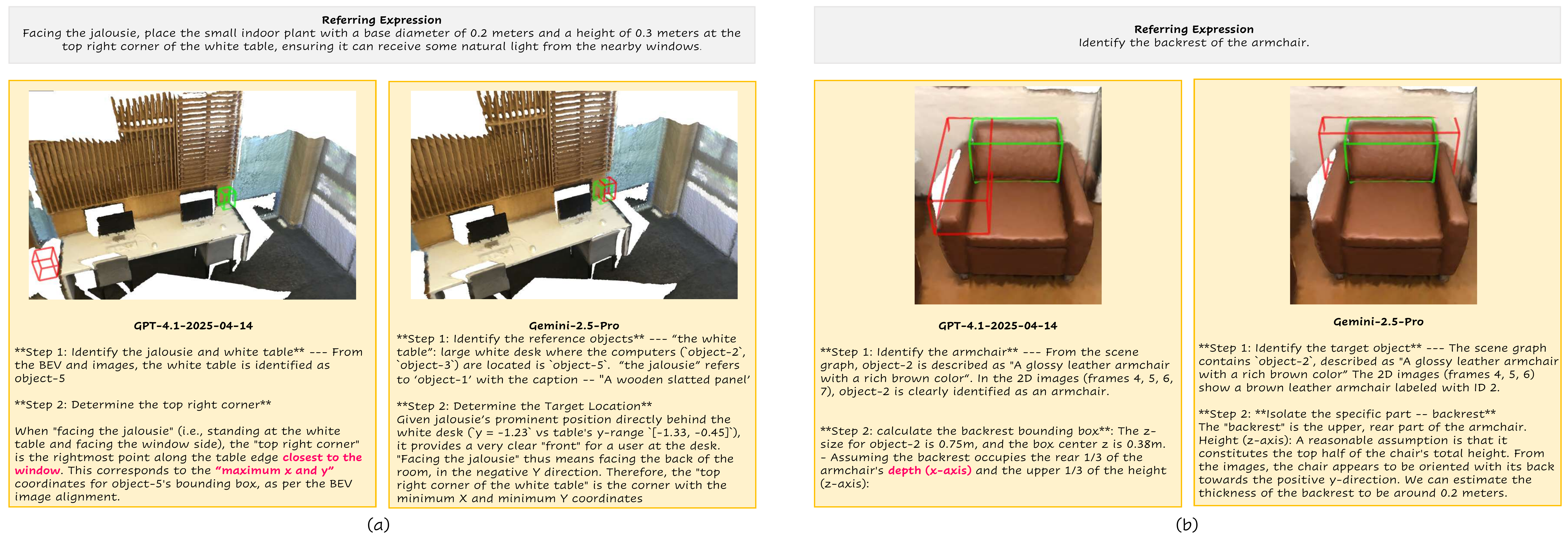}
    \caption{Qualitative results on GPT-4.1 and Gemini 2.5 pro. Green boxes indicate ground-truth, while red boxes show predictions from GPT-4.1 and Gemini-2.5-pro. The fatal error in reasoning process made by GPT-4.1 is highlighted in \textcolor{pink}{bold}.}
    \label{fig:Qualitative results fig2 on GPT-4.1 and Gemini 2.5 Pro}
\end{figure*}

In \cref{fig:Qualitative results fig2 on GPT-4.1 and Gemini 2.5 Pro} (a), which involves situational and relational understanding, Gemini-2.5-Pro correctly infers that ``facing the jalousie means facing the negative Y direction'' by analyzing the coordinates of the white table and the jalousie. In contrast, GPT-4.1 hallucinates an axis alignment that is not supported by the BEV image.

In \cref{fig:Qualitative results fig2 on GPT-4.1 and Gemini 2.5 Pro} (b), Gemini-2.5-Pro accurately reasons about the chair's orientation from the visual input, while GPT-4.1 fails to relate the depth of the armchair to the correct coordinate axes.

\subsection{More analysis on spatial reasoning}

As presented in \cref{fig:Prompt for Evaluating models' ability to transfer relative terms to axes.}, we present the prompt for imagine facing a given spatial axis (i.e., positive X, negative X, positive Y, or negative Y) and evaluate models' ability to correctly map the terms ``left'' and ``right'' to the corresponding spatial axes.

The right-handed coordinate system is commonly used in daily life and 3D modeling (also the same setting in Anywhere3D-Bench). For each model, we independently evaluate all eight scenarios: facing each of the four coordinate axes with queries about both the left and right sides. Each test is conducted three times, and we report the overall accuracy across 24 trials.

Experimental results in \cref{tab:GPT_4.1_Gemini_2.5_pro} show that reasoning models (o3, Gemini-2.5-Pro, DeepSeek-R1-671B) outperform non-reasoning models (GPT-4.1 and Qwen-2.5 series) in overall. Among the reasoning models, o3 achieves the best performance, followed by Gemini-2.5-Pro.

We then further study Gemini-2.5-pro's reasoning process under the case ``Facing negative X, which axes is in your right side?''. It provides several distinct reasoning approaches: (1) inferring by imagining itself within the scene and leveraging spatial imagination; (2) reasoning based on the definition of the right-handed coordinate system; and (3) reasoning using the cross product. However, all three approaches fail in providing the correct answer, revealing deficiencies in its spatial reasoning abilities and confusion in understanding spatial concepts.


The performance drop under the left-handed setting, which is an unfamiliar configuration, suggesting that even reasoning models rely partially on learned conventions under right-handed system rather than developing robust spatial understanding and reasoning ability. 

Moreover, We also tested a setting in which cardinal directions (north, south, east, and west) are used instead of coordinate axes(positive X, negative X, positive Y, negative Y), evaluating whether the model can correctly determine the left and right directions when facing each cardinal direction. 

The evaluation results under cardinal directions are much more satisfactory. Among all six models, only DeepSeek-R1-671B and Qwen-2.5-VL-72B make errors: DeepSeek-R1-671B answered incorrectly once out of 24 trials, while Qwen-2.5-VL-72B answered incorrectly in all three trials when asked for the right-hand direction while facing west. 

This suggests that these models may have encountered training data related to reasoning with cardinal directions (such as maps or navigation tasks), which accounts for their strong performance in this setting. In contrast, they are likely less exposed to spatial reasoning tasks involving left-handed or right-handed coordinate systems, resulting in poorer performance on those tasks.

\begin{figure}[htbp]
\centering
\begin{minipage}{\linewidth}\vspace{0mm}
    \centering
    \begin{tcolorbox}
    \fontsize{7.0pt}{0.8\baselineskip}\selectfont
    
    Suppose you are in a 3D scene with a right-handed XYZ coordinate system, where the XY plane represents the horizontal plane and the positive Z-axis points upward. If the direction you are facing is along the \blueprompt{facing\_direction} axis, what is the axes to your \blueprompt{side\_choice} side? (Please answer using positive X, negative X, positive Y, negative Y and explain the reason.)
    \end{tcolorbox}
    \caption{Prompt for Evaluating models' ability to transfer relative terms(``left'', ``right'') to axes where \blueprompt{facing\_direction} is replaced from following axes {``positive X'', ``negative X'', ``positive Y'', ``negative Y''} and \blueprompt{side\_choice} is replaced from following side {``left'', ``right''}.}
    \label{fig:Prompt for Evaluating models' ability to transfer relative terms to axes.}
\end{minipage}
\end{figure}

\subsection{MLLMs v.s. 3D visual grounding models}
Here we present the quantitative results of the best-performing 3D visual grounding models on Anywhere3D at the object level, as shown in the figure \cref{fig:chatscene_results} below. Figure (a), (b), and (c) illustrate failure cases of Chat-Scene on three object-level expressions regarding floor area, size and distance. The red bounding box represents the prediction of Chat-Scene, while the green bounding box denotes the ground-truth object. Notably, in these three cases, Gemini-2.5-Pro, the best-performing MLLM, predicts the correct answer. In Figure (a), Chat-Scene identifies the floor as an object with an area of 1 square meter, whereas the actual floor area is approximately 8.7 square meters. The ground-truth object is a table, with a diameter of about 1 meter. In Figure (b), Chat-Scene predicts that an object with a height of 0.4 meters is a chair, while the ground-truth object should be a trash bin. In Figure (c), the expression refers to a chair located less than half a meter away from the trash bin. Among all four chairs, only the ground-truth chair meets this criterion; the other three—including the one chosen by Chat-Scene—are all at least 1 meter away from the trash bin.

These examples demonstrate that even state-of-the-art 3D visual grounding models struggle with the quantitative estimation of object size and inter-object distance. Through our evaluation, we use ground-truth masks of 3D visual grounding models, and in quantitative examples above, the scene is also simple, requiring neither strong perception nor complex reasoning ability. Nonetheless, Chat-Scene still perform poorly on these quantitative object-level tasks. For space-level and part-level tasks, current 3D visual grounding models can only output object IDs; even if they could predict the corresponding bounding box centers and sizes, it is foreseeable that their performance would remain limited on tasks requiring logical computation and spatial reasoning.

\begin{figure}[ht]
  \centering
    \includegraphics[width=\linewidth]{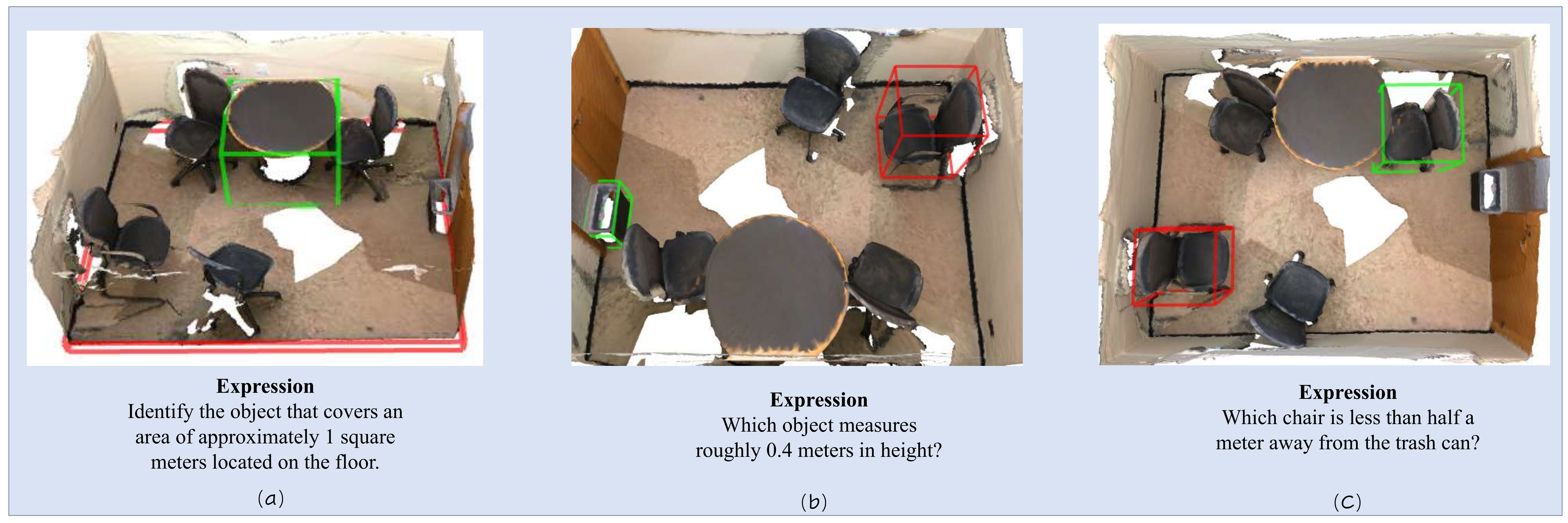}
    \caption{Qualitative results of Chat-Scene on object-level tasks. (a): Shape-related expression, (b): Size-related expression. (c): Distance-related expression. \textcolor{green}{Green boxes} indicate ground-truth, while \textcolor{red}{red boxes} show predictions from Chat-Scene. On these three cases, Gemini-2.5-pro, the best-performing MLLMs, predict the correct answers.}
    \label{fig:chatscene_results}
\end{figure}

\renewcommand\thefigure{D\arabic{figure}}
\setcounter{figure}{0}
\renewcommand\thetable{D\arabic{table}}
\setcounter{table}{0}
\renewcommand\theequation{D\arabic{equation}}
\setcounter{equation}{0}

\section{Detailed Related Work}
\label{appendix: detailed related work}

In this section, we provide a more detailed discussion of recent benchmarks related to spatial intelligence, including both task formatted as QA and visual grounding, and compare our proposed Anywhere3D-Bench with these benchmarks on four visual grounding levels.

\textbf{Object-level comparison with VSI-Bench:}
\begin{itemize}
    \item VSI-Bench is formulated as a \textit{question-answering} task, while Anywhere3D-Bench is a visual grounding benchmark.
    \item Besides, questions in VSI-Bench are constructed using eight \textbf{strictly formatted question templates} (see Table 4 in VSI-Bench's appendix). For example, object size questions in VSI-Bench are limited to a fixed form asking about ``the length of the longest dimension'', whereas Anywhere3D features more diverse queries about objects, such as objects' aspect ratio and occupied floor area (as exemplified in \cref{fig:data_stats}).
\end{itemize}

\textbf{Part-level comparison with SceneFun3D:}
\begin{itemize}
    \item Anywhere3D emphasizes the visual grounding of \textbf{part movements} by predicting the parts' positions after movement to test models' ability of moving parts in the 3D space (e.g., ``pull the top drawer out until it touches the armchair'' in \cref{fig:teaser}), whereas SceneFun3D is designed to predict the \textit{part’s original positions and motion directions}.
    \item In addition, SceneFun3D focuses on 9 affordance categories of functional interactive elements of objects, such as ``handles'' and ``knobs''. In contrast, Anywhere3D involves more \textbf{open-ended object parts} (e.g., ``toilet tank'', ``lampshade of the lamp'', ``top drawer of the cabinet'').
\end{itemize}

\textbf{Space- and region-level comparison with ScanReason/Space3D-Bench/MMScan:}

Quite different from the spatial questions in other benchmarks, the ``space-level'' queries in Anywhere3D are intended to ground \textbf{unoccupied space} (as explained in \cref{fig:teaser}'s caption and Abstract), often involving placing a new object or moving an existing object to a specified unoccupied space within the scene, such as:
\begin{itemize}
    \item \textit{``Place a cup on the upper right corner of the bedside table.''}(\cref{fig:teaser})
    \item \textit{``Mount a clock on the wall above the shelf for convenient viewing.''}(\cref{fig:Qualitative results fig1 on GPT-4.1 and Gemini 2.5 Pro})
    \item \textit{``Move the chair 0.5 meters backward.''}
\end{itemize}

Comparatively, ScanReason focuses on object-level visual grounding; Space3D-Bench focuses on question-answering tasks concerning objects and rooms, whereas the area-level queries in Anywhere3D-Bench are not limited to rooms, but also include \textbf{functional areas}, possibly a portion of a room, e.g., the study area in \cref{fig:teaser}. MMScan's region-level tasks are similar to our area-level tasks.

To clearly illustrate how Anywhere3D-Bench differs from prior benchmarks, we provide a detailed comparison in the \cref{tab: comparison spatial intelligence benchmark} below:

\begin{table}[h]
    \centering
    \caption{Comparison with recent benchmarks focusing on spatial intelligence}
    \label{tab: comparison spatial intelligence benchmark}
    \setlength{\tabcolsep}{10pt}
    \renewcommand{\arraystretch}{1.1}
    \resizebox{\linewidth}{!}{
    \begin{tabular}{cccccc}
    \toprule
    \textbf{Benchmark} & \textbf{Task Format} & \textbf{Area/Region/Room} & \textbf{Unoccupied Space} & \textbf{Object} & \textbf{Part}\\ 
    \midrule
   VSI-Bench     & template-based QA     & \cmark    & \xmark    & \cmark   & \xmark     \\
   SceneFun3D     & grounding     & \xmark     & \xmark     & \xmark    & \cmark(only 9 functional interactive classes)  \\ 
   MMScan & grounding + QA & \cmark & \xmark & \cmark &\xmark \\
   ScanReason & grounding & \xmark & \xmark & \cmark &\xmark \\
   Space3D-Bench & grounding  & \cmark & \xmark & \cmark &\xmark  \\
   \textbf{Anywhere3D-Bench(ours)} & grounding & \cmark & \cmark & \cmark &\cmark\\
    \bottomrule
    \end{tabular}
    }
\end{table}

As shown, Anywhere3D-Bench provides a holistic evaluation for \textbf{multi-level} grounding in 3D scenes, while other benchmarks touch only one or two levels. Also, the \textbf{space-level} tasks, requiring reasoning about unoccupied space beyond objects, represent a particularly novel aspect of our benchmark.

\section{Limitations and Future Directions}
\label{appendix: limitations and future directions}

In this section, we outline the limitations of our current work and propose directions for future research.

First, we plan to develop a training set for \textit{Anywhere3D-Bench}. This would enable the supervised fine-tuning or reinforcement learning-based fine-tuning strategies to enhance the multi-level visual grounding capabilities of both 3D visual grounding models and lightweight VLMs.

Second, we aim to perform a deeper analysis of the reasoning processes produced by models and design corresponding evaluation metrics to assess the correctness of intermediate reasoning steps. In some cases, even if the final bounding box prediction is correct, the intermediate reasoning may involve compensatory errors---e.g., two incorrect steps canceling each other out. To address this, we are hoping to construct dual-form expressions, such as converting ``Facing the sofa, place a table on the right side of the sofa'' into ``Imagine sitting on the sofa, place a table on the left side.'' A model's prediction should only be considered correct if it answers both dual expressions consistently.

Third, while current model outputs are represented as the center coordinates and size of a 3D bounding box, we are interested in exploring a multiple-choice formulation. In this setting, the model would select the correct bounding box from a set of candidates, including the ground-truth and several distractors sampled from the scene.

Fourth, the proposed visual perception and relational reasoning enhancements serve as initial attempts to improve performance on Anywhere3D-Bench, highlighting the substantial gap between current models and humans. We hope these efforts will inspire future explorations, such as adopting video sampling strategies or other advanced techniques.

Fifth, we hope to move beyond visual grounding alone and explore object generation at the grounded location, such as generating corresponding 2D images of objects placed in the predicted 3D position.

Sixth, many expressions in \textit{Anywhere3D-Bench} can naturally serve as instructions. In future work, we plan to extend the benchmark toward embodied tasks, enabling agents (e.g., robots or simulated avatars) to execute the instructions in interactive 3D environments.

\renewcommand\thefigure{E\arabic{figure}}
\setcounter{figure}{0}
\renewcommand\thetable{E\arabic{table}}
\setcounter{table}{0}
\renewcommand\theequation{E\arabic{equation}}
\setcounter{equation}{0}

\end{document}